\documentclass[11pt,onecolumn]{article}

\usepackage[applemac]{inputenc}
\usepackage[dvips]{graphics,graphicx,epsfig}

\usepackage{times}
\usepackage{latexsym}

\usepackage{amsmath} 
\usepackage{amssymb} 
\usepackage{stmaryrd}
\usepackage{verbatim}

\usepackage{pictex}
\usepackage{cite} 

\makeatletter
\def\@maketitle{%
  \vbox to 6.5cm{%
    \hsize\textwidth
    \linewidth\hsize
    \vspace{1.5cm}
    \centering
    {\bfseries\LARGE \@title \par}
    \vspace{36pt}
    {\fontsize{11pt}{13pt}\selectfont \begin{tabular}[t]{c}\@author \end{tabular}\par}
    \vfill} 
}

\renewcommand\section{\@startsection{section}{1}{\z@}%
                       {-12\p@ \@plus -4\p@ \@minus -4\p@}%
                       {6\p@ \@plus 4\p@ \@minus 4\p@}%
                       {\normalfont\large\bfseries
                        \rightskip=\z@ \@plus 8em\pretolerance=10000 }}
\renewcommand\subsection{\@startsection{subsection}{2}{\z@}%
                       {-12\p@ \@plus -4\p@ \@minus -4\p@}%
                       {6\p@ \@plus 4\p@ \@minus 4\p@}%
                       {\normalfont\fontsize{11pt}{13pt}\selectfont\bfseries
                        \rightskip=\z@ \@plus 8em\pretolerance=10000 }}
\renewcommand\subsubsection{\@startsection{subsubsection}{3}{\z@}%
                       {-12\p@ \@plus -4\p@ \@minus -4\p@}%
                       {6\p@ \@plus 4\p@ \@minus 4\p@}%
                       {\normalfont\normalsize\itshape}}
\renewcommand\paragraph{\@startsection{paragraph}{4}{\z@}%
                       {-12\p@ \@plus -4\p@ \@minus -4\p@}%
                       {-0.5em \@plus -0.22em \@minus -0.1em}%
                       {\normalfont\normalsize\itshape}}
\makeatother

\makeatletter
\def\blfootnote{\xdef\@thefnmark{}\@footnotetext}  
\makeatother

\renewenvironment{abstract}%
  {\small
    \list{}{\labelwidth0pt
      \leftmargin0pt \rightmargin\leftmargin
      \listparindent\parindent \itemindent0pt
      \parsep0pt
      }%
    \item[\hskip\labelsep\bfseries\abstractname\enspace --] \itshape}{\endlist}

\newcommand{\keywordsname}{Keywords}
\newenvironment{keywords}%
  {\small
    \list{}{\labelwidth0pt
      \leftmargin0pt \rightmargin\leftmargin
      \listparindent\parindent \itemindent0pt
      \parsep0pt
      }%
    \item[\hskip\labelsep\bfseries\keywordsname:]}{\endlist}

\begin{document}

\pagestyle{plain} 

\title{An introduction to DSmT}

\vspace{1cm}

\author{\begin{tabular}{c@{\extracolsep{8em}}c}
Jean~Dezert & Florentin~Smarandache\\
French Aerospace Research Lab., &  Chair of Math. \& Sciences Dept., \\
ONERA/DTIM/SIF, &  University of New Mexico, \\
29 Avenue de la Division Leclerc, & 200 College Road,\\
92320 Ch\^atillon, France. & Gallup, NM 87301, U.S.A.\\
{\tt jean.dezert@onera.fr} & {\tt smarand@unm.edu}
\end{tabular}}

\date{}
\maketitle
\pagestyle{plain}


\begin{abstract}
The management and combination of uncertain, imprecise, fuzzy and even paradoxical or high conflicting sources of information has always been, and still remains today, of primal importance for the development of reliable modern information systems involving artificial reasoning. In this introduction, we present a survey of our recent theory of plausible and paradoxical reasoning, known as Dezert-Smarandache Theory (DSmT), developed for dealing with imprecise, uncertain and conflicting sources of information. We focus our presentation on the foundations of DSmT and on its most important rules of combination, rather than on browsing specific applications of DSmT available in literature. Several simple examples are given throughout this presentation to show the efficiency and the generality of this new approach.
\end{abstract}



\blfootnote{This paper is based on the first chapter of  \cite{Ref-Book_2009}.}

\section{Introduction}

The management and combination of uncertain, imprecise, fuzzy and even paradoxical or high conflicting sources of information has always been, and still remains today, of primal importance for the development of reliable modern information systems involving artificial reasoning. The combination (fusion) of information arises in many fields of applications nowadays (especially in defense, medicine, finance, geo-science, economy, etc). When several sensors, observers or experts have to be combined together to solve a problem, or if one wants to update our current estimation of solutions for a given problem with some new information available, we need powerful and solid mathematical tools for the fusion, specially when the information one has to deal with is imprecise and uncertain. In this paper, we present a survey of our recent theory of plausible and paradoxical reasoning, known as Dezert-Smarandache Theory (DSmT) in the literature, developed for dealing with imprecise, uncertain and conflicting sources of information. Recent publications have shown the interest and the ability of DSmT to solve problems where other approaches fail, especially when conflict between sources becomes high. We focus this presentation rather on the foundations of DSmT, and on the main important rules of combination, than on browsing specific applications of DSmT available in literature. Several simple examples are given throughout the presentation to show the efficiency and the generality of DSmT.

\section{Foundations of DSmT}

The development of DSmT (Dezert-Smarandache Theory of plausible and paradoxical reasoning \cite{Ref-DSmTBook_2004a,Ref-Dezert_2006a}) arises from the necessity to overcome the inherent limitations of DST (Dempster-Shafer Theory \cite{Ref-Shafer_1976}) which are closely related with the acceptance of Shafer's model for the fusion problem under consideration (i.e. the frame of {\it{discernment}} $\Theta$ is implicitly defined as a finite set of {\it{exhaustive}} and {\it{exclusive}} hypotheses $\theta_i$, $i=1,\ldots,n$ since the masses of belief are defined only on the power set of $\Theta$ - see section \ref{Book3Chap1SecPowerset} for details), the third middle excluded principle (i.e. the existence of the complement for any elements/propositions belonging to the power set of $\Theta$), and the acceptance of Dempster's rule of combination (involving normalization) as the framework for the combination of independent sources of evidence. Discussions on limitations of DST and presentation of some alternative rules to Dempster's rule of combination can be found in \cite{Ref-Zadeh_1979,Ref-Zadeh_1984,Ref-Zadeh_1985,Ref-Yager_1985,Ref-Zadeh_1986,Ref-Dubois_1986c,Ref-Yager_1987,Ref-Pearl_1988,Ref-Smets_1988,Ref-Voorbraak_1991,Ref-Inagaki_1991,Ref-Murphy_2000,Ref-Lefevre_2002,Ref-Sentz_2002,Ref-Lefevre_2003,Ref-DSmTBook_2004a} and therefore they will be not reported in details in this introduction. We argue that these three fundamental conditions of DST can be removed and another new mathematical approach for combination of evidence is possible. This is the purpose of DSmT.\\

The basis of DSmT is  the refutation of the principle of the third excluded middle and Shafer's model, since for a wide class of fusion problems the intrinsic nature of hypotheses can be only vague and imprecise in such a way that precise refinement is just impossible to obtain in reality so that the exclusive elements $\theta_i$ cannot be properly identified and precisely separated. Many problems involving fuzzy continuous and relative concepts described in natural language and having no absolute interpretation like tallness/smallness, pleasure/pain, cold/hot, Sorites paradoxes, etc,  enter in this category. DSmT starts with the notion of {\it{free DSm model}}, denoted $\mathcal{M}^f(\Theta)$, and considers $\Theta$ only as a frame of exhaustive elements $\theta_i$, $i=1,\ldots,n$ which can potentially overlap. This model is {\it{free}} because no other assumption is done on the hypotheses, but the weak exhaustivity constraint which can always be satisfied according the closure principle explained in \cite{Ref-DSmTBook_2004a}. No other constraint is involved in the free DSm model.
When the free DSm model holds, the classic commutative and associative classical DSm rule of combination, denoted DSmC, corresponding to the conjunctive consensus defined on the free Dedekind's lattice is performed.\\

Depending on the intrinsic nature of the elements of the fusion problem under consideration, it can however happen that the free model does not fit the reality because some subsets of $\Theta$ can contain elements known to be truly exclusive but also truly non existing at all at a given time (specially when working on dynamic fusion problem where the frame $\Theta$ varies with time with the revision of the knowledge available). These integrity constraints are then explicitly and formally introduced into the free DSm model $\mathcal{M}^f(\Theta)$ in order to adapt it properly to fit as close as possible with the reality and permit to construct a {\it{hybrid DSm model}} $\mathcal{M}(\Theta)$ on which the combination will be efficiently performed. Shafer's model, denoted $\mathcal{M}^0(\Theta)$, corresponds to a very specific hybrid DSm model including all possible exclusivity constraints. DST has been developed for working only with $\mathcal{M}^0(\Theta)$ while DSmT has been developed for working with any kind of hybrid model (including Shafer's model and the free DSm model), to manage as efficiently and precisely as possible imprecise, uncertain and potentially high conflicting sources of evidence while keeping in mind the possible dynamicity of the information fusion problematic. The foundations of DSmT are therefore totally different from those of all existing approaches managing uncertainties, imprecisions and conflicts. DSmT provides a new interesting way to attack the  information fusion problematic with a general framework in order to cover a wide variety of problems. \\

DSmT refutes also the idea that sources of evidence provide their beliefs with the same absolute interpretation of elements of the same frame $\Theta$ and the conflict between sources arises not only because of the possible unreliability of sources, but also because of possible different and relative interpretation of $\Theta$, e.g. what is considered as good for somebody can be considered as bad for somebody else. There is some unavoidable subjectivity in the belief assignments provided by the sources of evidence, otherwise it would mean that all bodies of evidence have a same objective and universal interpretation (or measure) of the phenomena under consideration, which unfortunately rarely occurs in reality, but when basic belief assignments (bba's) are based on some {\it{objective probabilities}} transformations. But in this last case, probability theory can handle properly and efficiently the information, and DST, as well as DSmT, becomes useless. If we now get out of the probabilistic background argumentation for the construction of bba, we claim that in most of cases, the sources of evidence provide their beliefs about elements of the frame of the fusion problem only based on their own limited knowledge and experience without reference to the (inaccessible) absolute truth of the space of possibilities.  Several successful applications of DSmT (in target tracking, satellite surveillance, situation analysis, robotics, medicine, etc) can be found in \cite{Ref-DSmTBook_2004a,Ref-Book_2006}.

\subsection{The power set, hyper-power set and super-power set}
\label{Book3Chap1SecPowerset}
In DSmT, we take very care of the model associated with the set $\Theta$ of hypotheses where the solution of the problem is assumed to belong to. In particular, the three main sets (power set, hyper-power set and super-power set) can be used depending on their ability to fit adequately with the nature of hypotheses. In the following, we assume that $\Theta=\{\theta_{1},\ldots,\theta_{n}\}$ is a finite set (called frame) of $n$ exhaustive elements\footnote{We do not assume here that elements $\theta_i$ are necessary exclusive, unless specified. There is no restriction on $\theta_i$ but the exhaustivity.}. If $\Theta=\{\theta_{1},\ldots,\theta_{n}\}$ is a priori not closed ($\Theta$ is said to be an open world/frame), one can always include in it a closure element, say $\theta_{n+1}$ in such away that we can work with a new closed world/frame $\{\theta_{1},\ldots,\theta_{n},\theta_{n+1}\}$. So without loss of generality, we will always assume that we work in a closed world by considering the frame $\Theta$ as a finite set of exhaustive elements. Before introducing the power set, the hyper-power set and the super-power set it is necessary to recall that subsets are regarded as  propositions in Dempster-Shafer Theory (see Chapter 2 of \cite{Ref-Shafer_1976}) and we adopt the same approach in DSmT.

\begin{itemize}

\item {\textbf{Subsets as propositions}}: Glenn Shafer in pages 35--37 of \cite{Ref-Shafer_1976} considers the subsets as propositions in the case we are concerned with the true value of some quantity $\theta$ taking its possible values in $\Theta$. Then the propositions $ \mathcal{P}_{\theta}(A)$ of interest are those of the form\footnote{We use the symbol $\triangleq$ to mean \textit{equals by definition}; the right-hand side of the equation is the definition of the left-hand side.}:
$$ \mathcal{P}_{\theta}(A)\triangleq \text{{\textit{The true value of $\theta$ is in a subset $A$ of $\Theta$}}.}$$

Any proposition $\mathcal{P}_{\theta}(A)$ is thus in one-to-one correspondence with the subset $A$ of $\Theta$. Such correspondence is very useful since it translates the logical notions of conjunction $\wedge$, disjunction $\vee$, implication $\Rightarrow$ and negation $\neg$ into the set-theoretic notions of intersection $\cap$, union $\cup$, inclusion $\subset$ and complementation $c(.)$.
Indeed, if $\mathcal{P}_{\theta}(A)$ and $\mathcal{P}_{\theta}(B)$ are two propositions corresponding to subsets $A$ and $B$ of $\Theta$, then the conjunction $\mathcal{P}_{\theta}(A)\wedge \mathcal{P}_{\theta}(B)$ corresponds to the intersection $A\cap B$ and the disjunction $\mathcal{P}_{\theta}(A)\vee \mathcal{P}_{\theta}(B)$ corresponds to the union $A\cup B$. $A$ is a subset of $B$ if and only if $\mathcal{P}_{\theta}(A)\Rightarrow \mathcal{P}_{\theta}(B)$ and $A$ is the set-theoretic complement of $B$ with respect to $\Theta$ (written $A=c_{\Theta}(B)$) if and only if $\mathcal{P}_{\theta}(A)= \neg \mathcal{P}_{\theta}(B)$. In other words, the following equivalences are then used between the operations on the subsets and on the propositions:
 \begin{table}[!h]
\centering
 \begin{tabular}{|l|c|c|}
    \hline
            Operations  & Subsets & Propositions \\
    \hline
  Intersection/conjunction & $A\cap B$ & $\mathcal{P}_{\theta}(A)\wedge \mathcal{P}_{\theta}(B)$\\
  Union/disjunction & $A\cup B$ & $\mathcal{P}_{\theta}(A)\vee \mathcal{P}_{\theta}(B)$\\
 Inclusion/implication &  $A\subset B$ & $\mathcal{P}_{\theta}(A)\Rightarrow \mathcal{P}_{\theta}(B)$\\
 Complementation/negation & $A=c_{\Theta}(B)$ & $\mathcal{P}_{\theta}(A)= \neg 
  \mathcal{P}_{\theta}(B)$\\
    \hline
  \end{tabular}
  \caption{Correspondence between operations on subsets and on propositions.}
\label{Book3Chap1Table1}
\end{table}
\end{itemize}

\begin{itemize}
\item {\textbf{Canonical form of a proposition}}: In DSmT we consider all propositions/sets in a canonical form.  We take the disjunctive normal form, which is a disjunction of conjunctions, and it is unique in Boolean algebra and simplest. For example, $X = A\cap B\cap (A\cup B\cup C)$ it is not in a canonical form, but we simplify the formula and $X = A\cap B$ is in a canonical form.
\end{itemize}

\begin{itemize}
\item {\textbf{The power set}}: $2^\Theta\triangleq(\Theta,\cup)$
\end{itemize}
Aside Dempster's rule of combination, the power set is one of the corner stones of Dempster-Shafer Theory (DST) since the basic belief assignments to combine are defined on the power set of the frame $\Theta$. In mathematics, given a set $\Theta$, the power set of $\Theta$, written $2^\Theta$, is the set of all subsets of $\Theta$. In ZFC axiomatic set theory, the existence of the power set of any set is postulated by the axiom of power set. In other words, $\Theta$ generates the power set $2^\Theta$ with the $\cup$ (union) operator only.

\clearpage
\newpage

More precisely, the power set $2^\Theta$ is defined as the set of all composite propositions/subsets built from elements of $\Theta$ with $\cup$ operator such that: 
\begin{enumerate}
\item $\emptyset, \theta_1,\ldots, \theta_n \in 2^\Theta$.
\item  If $A, B \in 2^\Theta$, then $A\cup B\in 2^\Theta$.
\item No other elements belong to $2^\Theta$, except those obtained by using rules 1 and 2.
\end{enumerate}

\noindent
{\textbf{Examples of power sets}}:

\begin{itemize}
\item 
If $\Theta=\{\theta_{1},\theta_{2}\}$, then $2^{\Theta=\{\theta_{1},\theta_{2}\}}=\{\{\emptyset\},\{\theta_{1}\},\{\theta_{2}\},\{\theta_{1},\theta_{2}\}\}$ which is commonly written  as $2^\Theta=\{\emptyset,\theta_{1},\theta_{2},\theta_{1}\cup\theta_{2}\}$.
\end{itemize}

\begin{itemize}
\item 
Let's consider two frames $\Theta_{1}=\{A,B\}$ and $\Theta_{2}=\{X,Y\}$, then their power sets are  respectively $2^{\Theta_{1}=\{A,B\}}=\{\emptyset,A,B,A\cup B\}$ and $2^{\Theta_{2}=\{X,Y\}}=\{\emptyset,X,Y,X\cup Y\}$. Let's consider a refined frame $\Theta^{ref}=\{\theta_{1},\theta_{2},\theta_{3},\theta_{4}\}$. The granules $\theta_{i}$, $i=1,\ldots,4$ are not necessarily exhaustive, nor exclusive. If $A$ and $B$ are expressed more precisely in function of the granules $\theta_{i}$ by example as $A\triangleq \{\theta_{1},\theta_{2},\theta_{3}\}\equiv\theta_{1}\cup \theta_{2}\cup \theta_{3}$ and $B\triangleq \{\theta_{2},\theta_{4}\}\equiv \theta_{2}\cup \theta_{4}$ then the power sets can be expressed from the granules $\theta_{i}$ as follows:
\begin{align*}
2^{\Theta_{1}=\{A,B\}} & =\{\emptyset,A,B,A\cup B\}\\
& =\{ \emptyset, \underbrace{\{\theta_{1},\theta_{2},\theta_{3}\}}_{A}, 
\underbrace{\{\theta_{2},\theta_{4}\}}_{B},\underbrace{\{\{\theta_{1},\theta_{2},\theta_{3}\}, \{\theta_{2},\theta_{4}\} \}}_{A\cup B} \}\\
& = \{\emptyset,\theta_{1}\cup \theta_{2}\cup \theta_{3}, \theta_{2}\cup \theta_{4}, \theta_{1}\cup \theta_{2}\cup \theta_{3}\cup  \theta_{4}\}
\end{align*}

If $X$ and $Y$ are expressed more precisely in function of the finer granules $\theta_{i}$ by example as $X\triangleq \{\theta_{1}\}\equiv \theta_{1}$ and $Y\triangleq \{\theta_{2},\theta_{3},\theta_{4}\}\equiv \theta_{2}\cup\theta_{3}\cup \theta_{4}$ then: 
\begin{align*}
2^{\Theta_{2}=\{X,Y\}} & =\{\emptyset,X,Y,X\cup Y\}\\
& =\{ \emptyset, \underbrace{\{\theta_{1}\}}_{X}, 
\underbrace{\{\theta_{2},\theta_{3},\theta_{4}\}}_{Y},\underbrace{\{\{\theta_{1}\}, \{\theta_{2},\theta_{3},\theta_{4}\} \}}_{X\cup Y} \}\\
& = \{ \emptyset,\theta_{1}, \theta_{2}\cup\theta_{3}\cup \theta_{4}, \theta_{1}\cup \theta_{2}\cup \theta_{3}\cup  \theta_{4}\}
\end{align*}

\noindent
We see that one has naturally:
$$2^{\Theta_{1}=\{A,B\}}\neq  2^{\Theta_{2}=\{X,Y\}}\neq 2^{\Theta^{ref}=\{\theta_{1},\theta_{2},\theta_{3},\theta_{4}\}}$$
\noindent
even if working from $\theta_{i}$ with $A\cup B = X\cup Y =\{\theta_{1},\theta_{2},\theta_{3},\theta_{4}\} = \Theta^{ref}$.
\end{itemize}

\begin{itemize}
\item {\textbf{The hyper-power set}}: $D^\Theta\triangleq(\Theta,\cup,\cap)$
\end{itemize}

One of the cornerstones of DSmT is the free Dedekind's lattice \cite{Ref-Dedekind_1897} denoted {\it{hyper-power set}} in DSmT framework. Let $\Theta=\{\theta_{1},\ldots,\theta_{n}\}$ be a finite set (called frame) of $n$ exhaustive elements. The hyper-power set $D^\Theta$ is defined as the set of all composite propositions/subsets built from elements of $\Theta$ with $\cup$ and $\cap$ operators  such that: 
\begin{enumerate}
\item $\emptyset, \theta_1,\ldots, \theta_n \in D^\Theta$.
\item  If $A, B \in D^\Theta$, then $A\cap B\in D^\Theta$ and $A\cup B\in D^\Theta$.
\item No other elements belong to $D^\Theta$, except those obtained by using rules 1 or 2.
\end{enumerate}

Therefore by convention, we write $D^\Theta=(\Theta,\cup,\cap)$ which means that $\Theta$ generates $D^\Theta$ under operators $\cup$ and $\cap$.
The dual (obtained by switching $\cup$ and $\cap$ in expressions) of $D^\Theta$ is itself.  There are elements in $D^\Theta$ which are self-dual (dual to themselves), for example $\alpha_8$ for the case when $n=3$ in the following example. The cardinality of $D^\Theta$ is majored by 
$2^{2^n}$ when the cardinality of $\Theta$ equals $n$, i.e. $\vert\Theta\vert=n$. The generation 
of hyper-power set $D^\Theta$ is closely related with the famous Dedekind's problem \cite{Ref-Dedekind_1897,Ref-Comtet_1974} on enumerating the 
set of isotone Boolean functions. The generation of the hyper-power set is presented in \cite{Ref-DSmTBook_2004a}. Since for any given finite set $\Theta$, $\vert D^\Theta\vert  \geq \vert 2^\Theta\vert $ we call $D^\Theta$ the  {\it{hyper-power set}} of $\Theta$.\\

\noindent
{\textbf{Example of the first hyper-power sets}}:
\begin{itemize}
\item
For the degenerate case ($n=0)$ where $\Theta=\{  \}$, one has $D^\Theta=\{\alpha_0\triangleq\emptyset\}$ and $\vert D^\Theta\vert = 1$.
\item When $\Theta=\{\theta_{1}\}$, one has $D^\Theta=\{\alpha_0\triangleq\emptyset,\alpha_1\triangleq\theta_1 \}$ and $\vert D^\Theta\vert = 2$.
\item When $\Theta=\{\theta_{1},\theta_{2}\}$, one has
$D^\Theta=\{\alpha_0,\alpha_1,\ldots,\alpha_{4} \}$ and $\vert D^\Theta\vert = 5$ with
$\alpha_0\triangleq\emptyset$, $\alpha_1\triangleq\theta_1\cap\theta_2$, $\alpha_2\triangleq\theta_1$, $\alpha_3\triangleq\theta_2 $ and $\alpha_4\triangleq\theta_1\cup\theta_2 $.
\item
When $\Theta=\{\theta_{1},\theta_{2},\theta_{3}\}$,  one has $D^\Theta=\{\alpha_0,\alpha_1,\ldots,\alpha_{18} \}$ and $\vert D^\Theta\vert = 19$ with
\begin{equation*}
\begin{array}{ll}
\alpha_0\triangleq\emptyset          &                                       \\
\alpha_1\triangleq\theta_1\cap\theta_2\cap\theta_3    &\alpha_{10}\triangleq\theta_2   \\
\alpha_2\triangleq\theta_1\cap\theta_2    & \alpha_{11}\triangleq\theta_3                            \\
\alpha_3\triangleq\theta_1\cap\theta_3    &\alpha_{12}\triangleq(\theta_1\cap\theta_2)\cup\theta_3                            \\
\alpha_4\triangleq\theta_2\cap\theta_3    &\alpha_{13}\triangleq(\theta_1\cap\theta_3)\cup\theta_2                             \\
\alpha_5\triangleq(\theta_1\cup\theta_2)\cap\theta_3  & \alpha_{14}\triangleq(\theta_2\cap\theta_3)\cup\theta_1      \\
\alpha_6\triangleq(\theta_1\cup\theta_3)\cap\theta_2  & \alpha_{15}\triangleq\theta_1\cup\theta_2       \\
\alpha_7\triangleq(\theta_2\cup\theta_3)\cap\theta_1 & \alpha_{16}\triangleq\theta_1\cup\theta_3        \\
\alpha_8\triangleq(\theta_1\cap\theta_2)\cup(\theta_1\cap\theta_3)\cup(\theta_2\cap\theta_3)  & \alpha_{17}\triangleq\theta_2\cup\theta_3    \\
\alpha_9\triangleq\theta_1  & \alpha_{18}\triangleq\theta_1\cup\theta_2\cup\theta_3                                                   \end{array}
\end{equation*}
\end{itemize}

The cardinality of hyper-power set $D^\Theta$ for $n\geq1$ follows the sequence of Dedekind's numbers \cite{Ref-Sloane_2003}, i.e.
1,2,5,19,167, 7580,7828353,...  and analytical expression of Dedekind's numbers has been obtained recently by Tombak in \cite{Ref-Tombak_2001} (see \cite{Ref-DSmTBook_2004a} for details on generation and ordering of $D^\Theta$). Interesting investigations on the programming of the generation of hyper-power sets for engineering applications have been done in Chapter 15 of \cite{Ref-Book_2006}  and in \cite{Ref-Book_2009}.\\

\noindent
{\textbf{Examples of hyper-power sets}}:\\

Let's consider the frames $\Theta_{1}=\{A,B\}$ and $\Theta_{2}=\{X,Y\}$, then their corresponding hyper-power sets are $D^{\Theta_{1}=\{A,B\}}=\{\emptyset,A\cap B,A,B,A\cup B\}$ and $D^{\Theta_{2}=\{X,Y\}}=\{\emptyset,X\cap Y,X,Y,X\cup Y\}$. Let's consider a refined frame $\Theta^{ref}=\{\theta_{1},\theta_{2},\theta_{3},\theta_{4}\}$ where the granules $\theta_{i}$, $i=1,\ldots,4$ are now considered as {\textit{truly exhaustive and exclusive}}. If $A$ and $B$ are expressed more precisely in function of the granules $\theta_{i}$ by example as $A\triangleq \{\theta_{1},\theta_{2},\theta_{3}\}$ and $B\triangleq \{\theta_{2},\theta_{4}\}$ then 
\begin{align*}
D^{\Theta_{1}=\{A,B\}} &= \{\emptyset,A\cap B,A,B,A\cup B\}\\
& =\{ \emptyset, \underbrace{\{\theta_{1},\theta_{2},\theta_{3}\}\cap \{\theta_{2},\theta_{4}\} }_{A\cap B=\{\theta_{2}\}}, \underbrace{\{\theta_{1},\theta_{2},\theta_{3}\}}_{A}, 
\underbrace{\{\theta_{2},\theta_{4}\}}_{B},\\
& \qquad \qquad\qquad \qquad\qquad \qquad\qquad\underbrace{\{\{\theta_{1},\theta_{2},\theta_{3}\}, \{\theta_{2},\theta_{4}\} \}}_{A\cup B=\{\theta_{1},\theta_{2},\theta_{3},\theta_{4}\}} \}\\
& = \{ \emptyset, \theta_{2}, \theta_{1}\cup \theta_{2}\cup \theta_{3}, \theta_{2}\cup \theta_{4}, \theta_{1}\cup \theta_{2}\cup \theta_{3}\cup \theta_{4}\}\\
& \neq 2^{\Theta_{1}=\{A,B\}}
\end{align*}

If $X$ and $Y$ are expressed more precisely in function of the finer granules $\theta_{i}$ by example as $X\triangleq \{\theta_{1}\}$ and $Y\triangleq \{\theta_{2},\theta_{3},\theta_{4}\}$ then in assuming that $\theta_{i}$, $i=1,\ldots,4$ are exhaustive and exclusive, one gets

\begin{align*}
D^{\Theta_{2}=\{X,Y\}} & =\{\emptyset,X\cap Y,X,Y,X\cup Y\}\\
& =\{ \underbrace{\emptyset, \underbrace{\{\theta_{1}\}\cap \{\theta_{2},\theta_{3},\theta_{4}\} }_{X\cap Y=\emptyset}}_{\emptyset}, \underbrace{\{\theta_{1}\}}_{X}, 
\underbrace{\{\theta_{2},\theta_{3},\theta_{4}\}}_{Y},\underbrace{\{\{\theta_{1}\}, \{\theta_{2},\theta_{3},\theta_{4}\} \}}_{X\cup Y} \}\\
& =\{ \emptyset, \underbrace{\{\theta_{1}\}}_{X}, 
\underbrace{\{\theta_{2},\theta_{3},\theta_{4}\}}_{Y},\underbrace{\{\{\theta_{1}\}, \{\theta_{2},\theta_{3},\theta_{4}\} \}}_{X\cup Y} \}\\
& \equiv 2^{\Theta_{2}=\{X,Y\}}
\end{align*}

\noindent
Therefore, we see that $D^{\Theta_{2}=\{X,Y\}}\equiv 2^{\Theta_{2}=\{X,Y\}}$ because the exclusivity constraint $X\cap Y=\emptyset$ holds since one has assumed $X\triangleq \{\theta_{1}\}$ and $Y\triangleq \{\theta_{2},\theta_{3},\theta_{4}\}$ with exhaustive and exclusive granules $\theta_{i}$, $i=1,\ldots,4$. \\

If the granules $\theta_{i}$, $i=1,\ldots,4$ are not assumed exclusive, then of course the expressions of hyper-power sets cannot be simplified and one would have:

\begin{align*}
D^{\Theta_{1}=\{A,B\}} &= \{\emptyset,A\cap B,A,B,A\cup B\}\\
& =\{ \emptyset, (\theta_{1}\cup\theta_{2}\cup\theta_{3})\cap (\theta_{2}\cup \theta_{4}), \theta_{1}\cup \theta_{2}\cup \theta_{3}, \theta_{2}\cup \theta_{4}, \theta_{1}\cup \theta_{2}\cup \theta_{3}\cup \theta_{4}\}\\
& \neq 2^{\Theta_{1}=\{A,B\}}\\
& \\
D^{\Theta_{2}=\{X,Y\}} & =\{\emptyset,X\cap Y,X,Y,X\cup Y\}\\
& =\{\emptyset, \theta_{1}\cap (\theta_{2}\cup\theta_{3}\cup\theta_{4}), \theta_{1}, 
\theta_{2}\cup\theta_{3}\cup\theta_{4}, \theta_{1}\cup\theta_{2}\cup\theta_{3}\cup\theta_{4}\}\\
& \neq 2^{\Theta_{2}=\{X,Y\}}
\end{align*}

\noindent
{\textbf{Shafer's model of a frame}}: More generally, when all the elements of a given frame $\Theta$ are known (or are assumed to be) truly exclusive, then the hyper-power set $D^\Theta$ reduces to the classical power set $2^\Theta$. Therefore, working on power set $2^\Theta$ as Glenn Shafer has proposed in his Mathematical Theory of Evidence \cite{Ref-Shafer_1976}) is equivalent to work on hyper-power set $D^\Theta$ with the assumption that all elements of the frame are exclusive. 
This is what we call {\textit{Shafer's model of the frame}} $\Theta$, written $\mathcal{M}^0(\Theta)$, even if such model/assumption has not been clearly stated explicitly by Shafer himself in his milestone book.

\begin{itemize}
\item {\textbf{The super-power set}}: $S^\Theta\triangleq(\Theta,\cup,\cap,c(.))$
\end{itemize}

The notion of super-power set has been introduced by Smarandache in the Chapter 8 of \cite{Ref-Book_2006}. It corresponds actually to the theoretical construction of the power set of the minimal\footnote{The minimality refers here to the cardinality of the refined frames.} refined frame $\Theta^{ref}$ of $\Theta$. $\Theta$ generates $S^\Theta$ under operators $\cup$, $\cap$ and complementation $c(.)$. $S^\Theta=(\Theta,\cup,\cap,c(.))$ is a Boolean algebra with respect to the union, intersection and complementation. Therefore working with the super-power set is equivalent to work with a minimal theoretical refined frame $\Theta^{ref}$ satisfying Shafer's model. More precisely, $S^\Theta$ is defined as the set of all composite propositions/subsets built from elements of $\Theta$ with $\cup$, $\cap$ and $c(.)$ operators such that: 
\begin{enumerate}
\item $\emptyset, \theta_1,\ldots, \theta_n \in S^\Theta$.
\item  If $A, B \in S^\Theta$, then $A\cap B\in S^\Theta$, $A\cup B\in S^\Theta$.
\item  If $A \in S^\Theta$, then $c(A)\in S^\Theta$.
\item No other elements belong to $S^\Theta$, except those obtained by using rules 1, 2 and 3.
\end{enumerate}

As reported in \cite{Ref-SmarandacheUFT_2005}, a similar generalization has been previously used in 1993 by Guan and Bell \cite{Ref-GuanBell_1993} for the Dempster-Shafer rule using propositions in sequential logic and reintroduced in 1994 by Paris in his book \cite{Ref-Paris_1994}, page 4.\\

\noindent
{\textbf{Example of a super-power set}}:\\
\medskip

Let's consider the frame $\Theta=\{\theta_{1},\theta_{2}\}$ and let's assume $\theta_{1}\cap\theta_{2}\neq\emptyset$, i.e. $\theta_{1}$ and $\theta_{2}$ are not disjoint according to Fig. \ref{Book3Chap1FigFree2DDSmModel} where $A\triangleq p_{1}$ denotes the part of $\theta_{1}$ belonging only to $\theta_{1}$ ($p$ stands here for {\textit{part}}), $B\triangleq p_{2}$ denotes the part of $\theta_{2}$ belonging only to $\theta_{2}$ and $C\triangleq p_{12}$ denotes the part of $\theta_{1}$ and $\theta_{2}$ belonging to both. In this example, $S^{\Theta=\{\theta_{1},\theta_{2}\}}$ is then given by
$$S^{\Theta}=\{\emptyset,\theta_{1}\cap\theta_{2},\theta_{1},\theta_{2},\theta_{1}\cup\theta_{2},c(\emptyset), c(\theta_{1}\cap\theta_{2}), c(\theta_{1}), c(\theta_{2}), c(\theta_{1}\cup\theta_{2}) \}$$

\noindent
where $c(.)$ is the complement in $\Theta$. Since $c(\emptyset)=\theta_{1}\cup\theta_{2}$ and $c(\theta_{1}\cup\theta_{2})=\emptyset$, the super-power set is actually given by
$$S^{\Theta}=\{\emptyset,\theta_{1}\cap\theta_{2},\theta_{1},\theta_{2},\theta_{1}\cup\theta_{2}, c(\theta_{1}\cap\theta_{2}), c(\theta_{1}), c(\theta_{2})\}$$

Let's now consider the minimal refinement $\Theta^{ref}=\{A,B,C\}$ of $\Theta$ built by splitting the granules $\theta_{1}$ and $\theta_{2}$ depicted on the previous Venn diagram into disjoint parts (i.e. $\Theta^{ref}$ satisfies the Shafer's model) as follows:

\begin{figure}[!h]
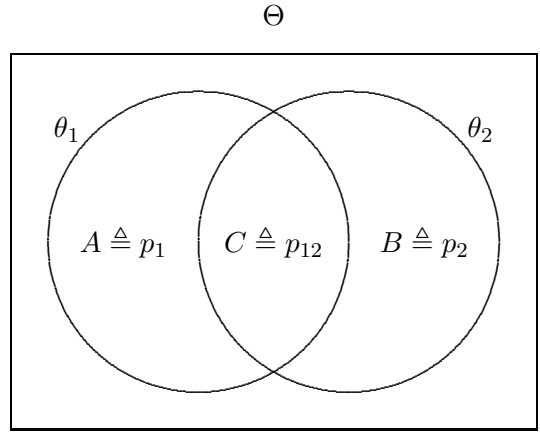

\begin{center}
\begin{minipage}{9cm}
\beginpicture
\setcoordinatesystem units <1cm,1cm> 
\setplotarea x from 1.5 to 8.5, y from 0.5 to 5.5 
\axis top label {$\Theta $} /
\axis bottom label {} /
\axis left label {} /
\axis right label {} /
\put {$\theta_{1}$} at 2.25 4.5 
\put {$A\triangleq p_{1}$} at 3 3 
\put {$\theta_{2}$} at 7.75 4.5 
\put {$B\triangleq  p_{2}$} at 7 3 
\put {$ C\triangleq p_{12}$} at 5 3 
\circulararc 360 degrees from 2 3 center at 4 3
\circulararc 360 degrees from 4 3 center at 6 3 
\endpicture
\end{minipage}
\end{center}
\caption{Venn diagram of a free DSm model for a 2D frame.}
\label{Book3Chap1FigFree2DDSmModel}
\end{figure}

$$\theta_{1}=A\cup C, \qquad \theta_{2}=B\cup C, \qquad \theta_{1}\cap\theta_{2}=C$$

\noindent
Then the classical power set of $\Theta^{ref}$ is given by

$$2^{\Theta^{ref}}=\{\emptyset,A,B,C,A\cup B,A\cup C,B\cup C, A\cup B\cup C\}$$

\noindent
We see that we can define easily a one-to-one correspondence, written $\sim$, between all the elements of the super-power set $S^{\Theta}$ and the elements of the power set $2^{\Theta^{ref}}$ as follows:
$$\emptyset \sim \emptyset, \quad (\theta_{1}\cap\theta_{2}) \sim C, \quad \theta_{1} \sim (A\cup C), \quad \theta_{2} \sim (B\cup C),\quad (\theta_{1}\cup \theta_{2}) \sim (A\cup B\cup C)$$
$$c(\theta_{1}\cap\theta_{2}) \sim (A\cup B),\quad c(\theta_{1}) \sim B, \quad c(\theta_{2}) \sim A$$

Such one-to-one correspondence between the elements of $S^{\Theta}$ and $2^{\Theta^{ref}}$ can be defined for any cardinality $|\Theta|\geq 2$ of the frame $\Theta$ and thus one can consider  $S^{\Theta}$ as the mathematical construction of the power set $2^{\Theta^{ref}}$ of the minimal refinement of the frame $\Theta$. Of course, when $\Theta$ already satisfies Shafer's model, the hyper-power set and the super-power set coincide with the classical power set of $\Theta$. It is worth to note that even if we have a mathematical tool to built the minimal refined frame satisfying Shafer's model, it doesn't mean necessary that one must work with this super-power set in general in real applications because most of the times the elements/granules of $S^{\Theta}$ have no clear physical meaning, not to mention the drastic increase of the complexity since one has $2^\Theta \subseteq D^\Theta \subseteq S^\Theta$ and
\begin{equation}
|2^\Theta|=2^{|\Theta |} < |D^\Theta|< |S^\Theta|=2^{^{|\Theta^{ref} |}}=2^{2^{|\Theta |}-1}
\end{equation}

\noindent
Typically,
\begin{table}[!h]
  \centering 
\begin{tabular}{|c|l|l|l|}
\hline
$\vert\Theta\vert=n$ & $\vert 2^\Theta \vert = 2^n$ & $\vert D^\Theta \vert$ & $\vert S^{\Theta} \vert = \vert 2^{\Theta_{ref}} \vert = 2^{2^n - 1}$ \\
\hline
2 &  4 & 5 &  $2^3=8$ \\
3 &  8 & 19 &  $2^7=128$ \\
4 &  16 & 167 &  $2^{15}=32768$ \\
5 &  32 & 7580 &  $2^{31}=2147483648$\\
\hline
\end{tabular}
\caption{Cardinalities of $2^\Theta$, $D^\Theta$ and $S^\Theta$. }
\label{Book3Chap1TabCardinalities}
\end{table}

In summary, DSmT offers truly the possibility to build and to work on refined frames and to deal with the complement whenever necessary, but in most of applications either the frame $\Theta$ is already built/chosen to satisfy Shafer's model or the refined granules have no clear physical meaning which finally prevent to be considered/assessed individually so that working on the hyper-power set is usually sufficient for dealing with uncertain imprecise (quantitative or qualitative) and highly conflicting sources of evidences. Working with $S^\Theta$ is actually very similar to working with $2^\Theta$ in the sense that in both cases we work with classical power sets; the only difference is that when working with $S^\Theta$ we have implicitly switched from the original frame $\Theta$ representation to a minimal refinement $\Theta^{ref}$ representation. Therefore, in the sequel we focus our discussions based mainly on hyper-power set rather than (super-) power set which has already been the basis for the development of DST. But as already mentioned, DSmT can easily deal with belief functions defined on $2^\Theta$ or $S^\Theta$ similarly as those defined on $D^\Theta$.\\

\noindent
{\textbf{Generic notation}}: In the sequel, we use the generic notation $G^\Theta$ for denoting the sets (power set, hyper-power set and super-power set) on which the belief functions are defined.\\

\begin{figure}[!h]
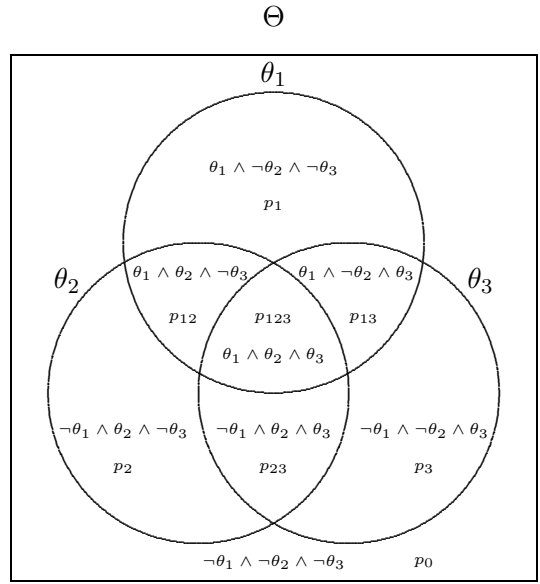

\begin{center}
\begin{minipage}{10cm}
\beginpicture
\setcoordinatesystem units <1cm,1cm>
\setplotarea x from 1.5 to 8.5, y from 0.5 to 7.5 
\axis bottom label {} /
\axis top label {$\Theta $} /
\axis left label {} /
\axis right label {} /
\put {$\theta_{1}$} at 5 7.25
\put {\tiny $\theta_{1} \wedge \neg \theta_{2} \wedge \neg \theta_{3}$} at 5 6 
\put {\tiny $p_{1}$} at 5 5.5
\put {$\theta_{3}$} at 7.75 4.5 
\put {$\theta_{2}$} at 2.25 4.5 
\put {\tiny $\neg \theta_{1} \wedge \theta_{2} \wedge \neg \theta_{3}$} at 3 2.5 
\put {\tiny $p_{2}$} at 3 2 
\put {\tiny $\neg \theta_{1} \wedge \neg \theta_{2} \wedge \theta_{3}$} at 7 2.5 
\put {\tiny $p_{3}$} at 7 2 
\put {\tiny $\neg \theta_{1} \wedge \theta_{2} \wedge \theta_{3}$} at 5 2.5 
\put {\tiny $p_{23}$} at 5 2
\put {\tiny $\theta_{1} \wedge \theta_{2} \wedge \theta_{3}$} at 5 3.5 
\put {\tiny $p_{123}$} at 5 4 
\put {\tiny $\theta_{1} \wedge \theta_{2} \wedge \neg \theta_{3}$} at 3.9 4.6 
\put {\tiny $p_{12}$} at 3.8 4
\put {\tiny $\theta_{1} \wedge \neg \theta_{2} \wedge \theta_{3}$} at 6.1 4.6 
\put {\tiny $p_{13}$} at 6.2 4 
\put {\tiny $\neg \theta_{1} \wedge \neg \theta_{2} \wedge \neg \theta_{3}$} at 5 .75 
\put {\tiny $p_{0}$} at 7 .75 
\circulararc 360 degrees from 3 5 center at 5 5
\circulararc 360 degrees from 2 3 center at 4 3
\circulararc 360 degrees from 4 3 center at 6 3 
\endpicture
\end{minipage}
\end{center}
\caption{Venn diagram of the free DSm model for a 3D frame.}
\label{Book3Chap1FigFree3DDSmModel}
\end{figure}

\noindent
{\textbf{Remark on the logical refinement}}: The refinement in logic theory presented recently by Cholvy in \cite{Ref-Cholvy2008} was actually proposed in nineties by a Guan and Bell \cite{Ref-GuanBell_1993} and by Paris \cite{Ref-Paris_1994}. This refinement  is isomorphic to the refinement in set theory done by many researchers.
If $\Theta = \{\theta_{1}, \theta_{2}, \theta_{3}\}$ is a language where the propositional variables are $\theta_{1}$, $\theta_{2}$, $\theta_{3}$, Cholvy considers all 8 possible logical combinations of propositions $\theta_{i}$'s or negations of  $\theta_{i}$'s (called interpretations), and defines the $8=2^3$ disjoint parts/propositions of the Venn diagram in Fig.~\ref{Book3Chap1FigFree3DDSmModel} [one also considers as a part the negation of the total ignorance] in the set theory, so that:
\begin{align*}
i_{1} &= \theta_{1}\wedge \theta_{2} \wedge \theta_{3}\\
i_{2} &= \theta_{1}\wedge\theta_{2} \wedge \neg\theta_{3}\\
i_{3} &= \theta_{1}\wedge  \neg\theta_{2} \wedge \theta_{3}\\
i_{4} &= \theta_{1}\wedge  \neg\theta_{2} \wedge \neg\theta_{3}\\
i_{5} &= \neg\theta_{1}\wedge \theta_{2} \wedge  \wedge \theta_{3}\\
i_{6} &= \neg\theta_{1}\wedge \theta_{2}  \wedge \neg\theta_{3}\\
i_{7} &= \neg\theta_{1}\wedge  \neg\theta_{2} \wedge \theta_{3}\\
i_{8} &= \neg\theta_{1}\wedge \neg\theta_{2}\wedge\neg\theta_{3}
\end{align*}
\noindent where $\neg\theta_{i}$ means the negation of $\theta_{i}$.

\clearpage
\newpage

Because of Shafer's equivalence of subsets and propositions, Cholvy's logical refinement is strictly equivalent to the refinement we did already in 2006 in defining $S^\Theta$ - see Chap. 8 of  \cite{Ref-Book_2006} -  but in the set theory framework. We did it using Smarandache’s codification (easy to understand and read) in the following way:
\begin{enumerate}
\item[-] each Venn diagram disjoint part $p_{ij}$, or $p_{ijk}$ represents respectively the intersection of $p_i$ and $p_j$ only, or $p_i$ and $p_j$ and $p_k$ only, etc; while the complement of the total ignorance is considered $p_0$ [$p$ stands for part].
\end{enumerate}

Thus, we have an easier and clearer representation in DSmT than in Cholvy’s logical representation. While the refinement in DST using logical approach for $n$ very large is very hard, we can simply consider in the DSmT the super-power set $S^\Theta = (\Theta, \cup, \cap, c(.))$. So, in DSmT representation the disjoint parts are noted as follows:
\begin{align*}
p_{123} & = \theta_{1}\wedge \theta_{2} \wedge \theta_{3}         = i_{1}\\
p_{12}   & = \theta_{1}\wedge\theta_{2} \wedge \neg\theta_{3} =  i_{2}\\
p_{13}   &= \theta_{1}\wedge  \neg\theta_{2} \wedge \theta_{3}  = i_{3}\\
p_{23}   & = \neg\theta_{1}\wedge \theta_{2} \wedge \theta_{3} =  i_{5}\\
p_{1} &= \theta_{1}\wedge  \neg\theta_{2} \wedge \neg\theta_{3} = i_{4}\\
p_{2} &= \neg\theta_{1}\wedge \theta_{2}  \wedge \neg\theta_{3} = i_{6}\\
p_{3} &= \neg\theta_{1}\wedge  \neg\theta_{2} \wedge \theta_{3} = i_{7}\\
p_{0} & = \neg\theta_{1}\wedge \neg\theta_{2}\wedge\neg\theta_{3} = i_{8}
\end{align*}

As seeing, in Smarandache’s codification a disjoint Venn diagram part is equal to the intersection of singletons whose indexes show up as indexes of the Venn part; for example in $p_{12}$ case indexes 1 and 2, intersected with the complement of the missing indexes, in this case index 3 is missing.\\

\noindent
Smarandache’s codification can easily transform any set from $S^\Theta$ into its canonical disjunctive normal form.  For example,  $\theta_{1} = p_{1} \cup  p_{12} \cup p_{13} \cup p_{123}$ (i.e. all Venn diagram disjoint parts that contain the index “1” in their indexes ; such indexes from $S^\Theta$ are 1, 12, 13, 123) can be expressed as
$$\theta_{1} = (\theta_{1} \cap  c(\theta_{2}) \cap c(\theta_{3})) \cup (\theta_{1}  \cap \theta_{2} \cap  c(\theta_{3})) \/ (\theta_{1}  \cap c(\theta_{2}) \cap \theta_{3}) \cup (\theta_{1}  \cap \theta_{2} \cap \theta_{3})$$
\noindent
 where the set values of each part was taken from the above table.

$\theta_{1}\wedge \theta_{2} = p_{12} \cup p_{123}$ (i.e. all Venn diagram disjoint parts that contain the index “12” in their indexes) equals to $(\theta_{1}\wedge\theta_{2} \wedge \neg\theta_{3}) \vee (\theta_{1}\wedge \theta_{2} \wedge \theta_{3})$.\\
 
The refinement based on Venn Diagram, becomes very hard and almost impossible when the cardinal of $\Theta$, $n$, is large and all intersections are non-empty (the free model). Suppose $n = 20$, or even bigger, and we have the free model.  How can we construct a Venn Diagram where to show all possible intersections of 20 sets? Its geometrical figure would be very hard to design and very hard to read (you don't identify well each disjoint part of a such Venn Diagram to what intersection of sets it belongs to). The larger is $n$, the more difficult is the refinement. Fortunately, based on Smarandache's codification, we can algebraically design in an easy way for all such intersections (for example, if $n$ is very big, we can use computer programs to make combinations of indexes $\{1,2,...,n\}$ taken in groups or 1, of 2, ..., or of $n$ elements each), so the refinement should not be a big problem from the programming point of view, but we must always keep in mind if such refinement is really necessary and if it has (or not) a deep physical interpretation and justification for the problem under consideration.\\

The assertion in \cite{Ref-Cholvy2008}, upon Milan Daniel’s, that hybrid DSm rule is equivalent to Dubois-Prade rule is untrue, since in dynamic fusion they give different results.  Such example has been already given in \cite{Ref-DezertPanel2004} and is reported in section \ref{Ref-Comparison} for the sake of clarification for the readers. The assertion in  \cite{Ref-Cholvy2008} that “from an expressivity point of view DSmT is equivalent to DST” is partially true since this idea is true when the refinement is possible (not always it is practically/physically possible), and even when the spaces we work on, $S^\Theta = 2^{\Theta^{ref}}$, where the hypotheses are exclusive, DSmT offers the advantage that the refinement is already done (it is not necessary for the user to do (or implicitly presuppose) it as in DST). Also, DSmT accepts from the very beginning the possibility to deal with non-exclusive hypotheses and of course it can a fortiori deal with sets of exclusive hypothesis and work either on $2^{\Theta}$ or $2^{\Theta^{ref}}$ whenever necessary, while DST first requires implicitly to work with exclusive hypotheses only.\\

The main distinctions between DSmT and DST are summarized by the following points:
\begin{enumerate}
\item The refinement is not always (physically) possible, especially for elements from the frame of discernment whose frontiers are not clear, such as: colors, vague sets, unclear hypotheses, etc. in the frame of discernment; DST does not fit well for working in such cases, while DSmT does;
\item  Even in the case when the frame of discernment can be refined (i.e. the {\textit{atomic}} elements of the frame have all a distinct physical meaning), it is still easier to use DSmT than DST since in DSmT framework the refinement is done automatically by the mathematical construction of the super-power set;
\item  DSmT offers better fusion rules, for example Proportional Conflict redistribution Rule \# 5 (PCR5) - presented in the sequel -  is better than Dempster's rule; hybrid DSm rule (DSmH) works for the dynamic fusion, while Dubois-Prade fusion rule does not (DSmH is an extension of Dubois-Prade rule);
\item DSmT offers the best qualitative operators (when working with labels) giving the most accurate and coherent results;
\item DSmT offers new interesting quantitative conditioning rules (BCRs) and qualitative conditioning rules (QBCRs), different from Shafer's conditioning rule (SCR). SCR can be seen simply as a combination of a prior mass of belief with the mass $m(A)=1$ whenever $A$ is the conditioning event;
\item DSmT proposes a new approach for working with imprecise quantitative or qualitative information and not limited to interval-valued belief structures as proposed generally in the literature \cite{Ref-Denoeux_1997,Ref-Denoeux_1999,Ref-Wang_2007}. 
\end{enumerate}

\subsection{Notion of free and hybrid DSm models}
\label{Chapter1Sec:DSMmodels}

{\textbf{Free DSm model}}: The elements $\theta_i$, $i=1,\ldots,n$ of $\Theta$ constitute the finite set of hypotheses/concepts characterizing the fusion problem under consideration. When there is no constraint on the elements of the frame, we call this model the {\textit{free DSm model}}, written $\mathcal{M}^f(\Theta)$. This free DSm model allows to deal directly with fuzzy concepts which depict a continuous and relative intrinsic nature and which cannot be precisely refined into finer disjoint information granules having an absolute interpretation because of the unreachable universal truth. In such case, the use of the hyper-power set $D^\Theta$ (without integrity constraints) is particularly well adapted for defining the belief functions one wants to combine.\\

\noindent
{\textbf{Shafer's model}}: In some fusion problems involving discrete concepts, all the elements $\theta_i$, $i=1,\ldots,n$ of $\Theta$ can be truly exclusive. In such case, all the exclusivity constraints on $\theta_i$, $i=1,\ldots,n$ have to be included in the previous model to characterize properly the true nature of the fusion problem and to fit it with the reality. 
By doing this, the hyper-power set $D^\Theta$ as well as the super-power set $S^\Theta$ reduce naturally to the classical power set $2^\Theta$ and this constitutes what we have called {\textit{Shafer's model}}, denoted $\mathcal{M}^0(\Theta)$. Shafer's model corresponds actually to the most restricted hybrid DSm model.\\

\noindent
{\textbf{Hybrid DSm models}}: Between the class of fusion problems corresponding to the free DSm model $\mathcal{M}^f(\Theta)$ and  the class of fusion problems corresponding to Shafer's model $\mathcal{M}^0(\Theta)$, there exists another wide class of hybrid fusion problems involving in $\Theta$ both fuzzy continuous concepts and discrete hypotheses. In such (hybrid) class, some exclusivity constraints and possibly some non-existential constraints (especially when working on dynamic\footnote{i.e. when the frame $\Theta$ and/or the model $\mathcal{M}$ is changing with time.} fusion) have to be taken into account. Each hybrid fusion problem of this class will then be characterized by a proper hybrid DSm model denoted $\mathcal{M}(\Theta)$ with $\mathcal{M}(\Theta)\neq\mathcal{M}^f(\Theta)$ and $\mathcal{M}(\Theta)\neq \mathcal{M}^0(\Theta)$. \\

In any fusion problems, we consider as primordial at the very beginning and before combining information expressed as belief functions to define clearly the proper frame $\Theta$ of the given problem and to choose explicitly its corresponding model one wants to work with. Once this is done, the second important point is to select the proper set $2^\Theta$, $D^\Theta$ or $S^\Theta$ on which the belief functions will be defined. The third important point will be the choice of an efficient rule of combination of belief functions and finally the criteria adopted for decision-making.\\

In the sequel, we focus our presentation mainly on hyper-power set $D^\Theta$ (unless specified) since it the most interesting new aspect of DSmT for readers already familiar with DST framework, but a fortiori we can work similarly on classical power set $2^\Theta$ if Shafer's model holds, and even on $2^{\Theta^{ref}}$ (the power set of the minimal refined frame) whenever one wants to use it and if possible.\\

\noindent
{\textbf{Examples of models for a frame $\Theta$}}: \\

\noindent $\bullet$ Let's consider the 2D problem where $\Theta=\{\theta_1,\theta_2\}$ with $D^\Theta=\{\emptyset,\theta_1\cap\theta_2,\theta_1,\theta_2,\theta_1\cup\theta_2\}$ and assume now that $\theta_1$ and $\theta_2$ are truly exclusive (i.e. Shafer's model $\mathcal{M}^0$ holds), then because $\theta_1\cap\theta_2\overset{\mathcal{M}^0}{=}\emptyset$, one gets $D^\Theta=\{\emptyset,\theta_1\cap\theta_2\overset{\mathcal{M}^0}{=}\emptyset,\theta_1,\theta_2,\theta_1\cup\theta_2\}=\{\emptyset,\theta_1,\theta_2,\theta_1\cup\theta_2\}\equiv 2^\Theta$.\\

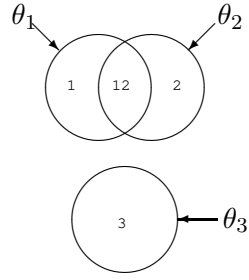
\begin{figure}[!h]
\begin{center}
{\tt \setlength{\unitlength}{1pt}
\begin{picture}(90,90)
\thinlines    
\put(40,60){\circle{40}}
\put(60,60){\circle{40}}
\put(50,10){\circle{40}}
\put(15,84){\vector(1,-1){10}}
\put(7,84){$\theta_{1}$}
\put(84,84){\vector(-1,-1){10}}
\put(85,84){$\theta_{2}$}
\put(85,10){\vector(-1,0){15}}
\put(87,7){$\theta_{3}$}
\put(45,59){\tiny{12}}
\put(47,7){\tiny{3}}
\put(68,59){\tiny{2}}
\put(28,59){\tiny{1}}
\end{picture}}
\end{center}
\caption{Venn diagram of a DSm hybrid model for a 3D frame.}
\label{Book3Chap1FigHybrid3DDSmModel}
\end{figure}

\noindent $\bullet$ As another simple example of hybrid DSm model, let's consider the 3D case with the frame $\Theta=\{\theta_1,\theta_2,\theta_3\}$ with the model $\mathcal{M}\neq\mathcal{M}^f$ in which we force all possible conjunctions to be empty, but $\theta_1\cap\theta_2$. This hybrid DSm model is then represented with  the Venn diagram on Fig. \ref{Book3Chap1FigHybrid3DDSmModel} (where boundaries of intersection of $\theta_1$ and $\theta_2$ are not precisely defined if $\theta_1$ and $\theta_2$ represent only fuzzy concepts like {\it{smallness}} and {\it{tallness}} by example).

\subsection{Generalized belief functions}

From a general frame $\Theta$, we define a map $m(.): G^\Theta \rightarrow [0,1]$ associated to a given body of evidence $\mathcal{B}$ as 
\begin{equation}
m(\emptyset)=0 \qquad \text{and}\qquad \sum_{A\in G^\Theta} m(A) = 1 
\end{equation}
\noindent The quantity $m(A)$ is called the {\it{generalized basic belief assignment/mass}} (gbba) of $A$.\\

\noindent
The {\it{generalized belief and plausibility functions}} are defined in almost the same manner as within DST, i.e.
\begin{equation}
\text{Bel}(A) = \sum_{\substack{B\subseteq A\\ B\in G^\Theta}} m(B)
\qquad\qquad\text{Pl}(A) = \sum_{\substack{B\cap A\neq\emptyset \\ B\in G^\Theta}} m(B)
\end{equation}

We recall that $G^\Theta$ is the generic notation for the set on which the gbba is defined ($G^\Theta$ can be  $2^\Theta$,  $D^\Theta$ or even  $S^\Theta$ depending on the model chosen for $\Theta$). These definitions are compatible with the definitions of the classical belief functions  in DST framework when $G^\Theta=2^\Theta$ for fusion problems where Shafer's model $\mathcal{M}^0(\Theta)$ holds. We still have $\forall A\in G^\Theta,\, \text{Bel}(A)\leq \text{Pl}(A)$. Note that when working with the free DSm model $\mathcal{M}^f(\Theta)$, one has always $\text{Pl}(A) =1$ $\forall A\neq\emptyset \in (G^\Theta=D^\Theta) $ which is normal.\\

\noindent
{\textbf{Example}}: Let's consider the simple frame $\Theta = \{A, B\}$, then depending on the model we choose for $G^{\Theta}$, one will consider either:
\begin{itemize}
\item  $G^{\Theta}$ as the power set  $2^\Theta$ and therefore:
$$m(A)+m(B)+m(A\cup B)=1$$
\item  $G^{\Theta}$ as the hyper-power set  $D^\Theta$ and therefore:
$$m(A)+m(B)+m(A\cup B) + m(A\cap B)=1$$
\item  $G^{\Theta}$ as the super-power set  $S^\Theta$ and therefore:
$$m(A)+m(B)+m(A\cup B) + m(A\cap B) + m(c(A)) + m(c(B))+ m(c(A)\cup c(B))=1$$
\end{itemize}

\subsection{The classic DSm rule of combination}

When the free DSm model $\mathcal{M}^f(\Theta)$ holds for the fusion problem under consideration, the  classic DSm rule of combination $m_{\mathcal{M}^f(\Theta)}\equiv m(.)\triangleq [m_{1}\oplus m_{2}](.)$ of two independent\footnote{While independence is a difficult concept to define in all theories managing epistemic uncertainty, we follow here the interpretation of Smets in \cite{Ref-Smets_1986} and \cite{Ref-Smets_1988}, p. 285 and consider that two sources of evidence are independent (i.e distinct and noninteracting) if each leaves one totally ignorant about the particular value the other will take.} sources of evidences $\mathcal{B}_{1}$ and  $\mathcal{B}_{2}$ over the same 
frame $\Theta$ with belief functions $\text{Bel}_{1}(.)$ and 
 $\text{Bel}_{2}(.)$ associated with gbba $m_{1}(.)$ and $m_{2}(.)$ corresponds to the conjunctive consensus of the sources. It is  given by \cite{Ref-DSmTBook_2004a}:
 \begin{equation}
\forall C\in D^\Theta,\qquad m_{\mathcal{M}^f(\Theta)}(C) \equiv m(C) = 
 \sum_{\substack{A,B\in D^\Theta\\ A\cap B=C}}m_{1}(A)m_{2}(B)
 \label{Chapter1JDZT}
 \end{equation}
 
Since $D^\Theta$ is closed under $\cup$ and $\cap$ set operators, this new rule 
of combination guarantees that $m(.)$ is a proper generalized belief assignment, i.e. $m(.): D^\Theta \rightarrow [0,1]$. This rule of combination is commutative and associative 
and can always be used for the fusion of sources involving fuzzy concepts when free DSm model holds for the problem under consideration. This rule has been extended for $s > 2$ sources in \cite{Ref-DSmTBook_2004a}.

\clearpage
\newpage

According to Table \ref{Book3Chap1TabCardinalities}, this classic DSm rule of combination looks very expensive in terms of computations and memory size due to the huge number of elements in $D^\Theta$ when the cardinality of $\Theta$ increases. This remark is however valid only if the cores (the set of focal elements of gbba) $\mathcal{K}_1(m_1)$ and $\mathcal{K}_2(m_2)$ coincide with $D^\Theta$, i.e. when $m_1(A)>0$ and $m_2(A)>0$ for all $A\neq\emptyset\in D^\Theta$. Fortunately, it is important to note here that in most of the practical applications the sizes of $\mathcal{K}_1(m_1)$ and $\mathcal{K}_2(m_2)$ are much smaller than $\vert D^\Theta\vert$ because bodies of evidence generally allocate their basic belief assignments only over a subset of the hyper-power set. This makes things easier for the implementation of the classic DSm rule \eqref{Chapter1JDZT}. The DSm rule is actually very easy to implement. It suffices for each focal element of $\mathcal{K}_1(m_1)$ to multiply it with the focal elements of $\mathcal{K}_2(m_2)$ and then to  pool all combinations which are equivalent under the  algebra of sets. While very costly in term on memory storage in the worst case (i.e. when all $m(A)>0$, $A\in D^\Theta$ or $A\in2^{\Theta^{ref}}$), the DSm rule however requires much smaller memory storage than when working with $S^\Theta$, i.e. working with a minimal refined frame satisfying Shafer's model.\\
 
In most fusion applications only a small subset of elements of $D^\Theta$ have a non null basic belief mass because all the commitments are just usually impossible to obtain precisely when the dimension of the problem increases. Thus, it is not necessary to generate and keep in memory all elements of $D^\Theta$ (or eventually $S^\Theta$) but only those which have a positive belief mass.  However there is a real technical challenge on how to manage efficiently all elements of the hyper-power set. This problem is obviously much more difficult when trying to work on a refined frame of discernment $\Theta^{ref}$ if one really prefers to use Dempster-Shafer theory and apply Dempster's rule of combination. It is important to keep in mind that the ultimate and minimal refined frame consisting in exhaustive and exclusive finite set of refined exclusive hypotheses is just impossible to justify and to define precisely for all problems dealing with fuzzy and ill-defined continuous concepts. A discussion on refinement with an example has be included in \cite{Ref-DSmTBook_2004a}.

\subsection{The hybrid DSm rule of combination}

When the free DSm model $\mathcal{M}^f(\Theta)$ does not hold due to the true nature of the fusion problem under consideration which requires to take into account some known integrity constraints, one has to work with a proper hybrid DSm model $\mathcal{M}(\Theta)\neq\mathcal{M}^f(\Theta)$. In such case, the hybrid DSm rule (DSmH) of combination based on the chosen hybrid DSm model $\mathcal{M}(\Theta)$ for $k\geq 2$ independent sources of information is defined for all $A\in D^\Theta$ as \cite{Ref-DSmTBook_2004a}:
\begin{equation}
m_{DSmH}(A)=m_{\mathcal{M}(\Theta)}(A)\triangleq 
\phi(A)\Bigl[ S_1(A) + S_2(A) + S_3(A)\Bigr]
 \label{Chapter1eq:DSmHkBis1}
\end{equation}
\noindent
where all sets involved in formulas are in the canonical form and $\phi(A)$ is the {\it{characteristic non-emptiness function}} of a set $A$, i.e. $\phi(A)= 1$ if  $A\notin \boldsymbol{\emptyset}$ and $\phi(A)= 0$ otherwise, where $\boldsymbol{\emptyset}\triangleq\{\boldsymbol{\emptyset}_{\mathcal{M}},\emptyset\}$. $\boldsymbol{\emptyset}_{\mathcal{M}}$ is the set  of all elements of $D^\Theta$ which have been forced to be empty through the constraints of the model $\mathcal{M}$ and $\emptyset$ is the classical/universal empty set. $S_1(A)\equiv m_{\mathcal{M}^f(\theta)}(A)$, $S_2(A)$, $S_3(A)$ are defined by 
\begin{equation}
S_1(A)\triangleq \sum_{\substack{X_1,X_2,\ldots,X_k\in D^\Theta\\ X_1\cap X_2\cap\ldots\cap X_k=A}} \prod_{i=1}^{k} m_i(X_i)
\end{equation}
\begin{equation}
S_2(A)\triangleq \sum_{\substack{X_1,X_2,\ldots,X_k\in\boldsymbol{\emptyset}\\ [\mathcal{U}=A]\vee [(\mathcal{U}\in\boldsymbol{\emptyset}) \wedge (A=I_t)]}} \prod_{i=1}^{k} m_i(X_i)
\end{equation}

\begin{equation}
S_3(A)\triangleq\sum_{\substack{X_1,X_2,\ldots,X_k\in D^\Theta \\ X_1\cup X_2\cup\ldots\cup X_k=A \\ X_1\cap X_2\cap \ldots\cap X_k \in\boldsymbol{\emptyset}}}  \prod_{i=1}^{k} m_i(X_i)
\end{equation}
with $\mathcal{U}\triangleq u(X_1)\cup u(X_2)\cup \ldots \cup u(X_k)$ where $u(X)$ is the union of all $\theta_i$ that compose $X$, $I_t \triangleq \theta_1\cup \theta_2\cup\ldots\cup \theta_n$ is the total ignorance.
$S_1(A)$ corresponds to the classic DSm rule for $k$ independent sources based on the free DSm model $\mathcal{M}^f(\Theta)$; $S_2(A)$ represents the mass of all relatively and absolutely empty sets which is transferred to the total or relative ignorances associated with non existential constraints (if any, like in some dynamic problems); $S_3(A)$ transfers the sum of relatively empty sets directly onto the canonical disjunctive form of non-empty sets.\\

The hybrid DSm rule of combination generalizes the classic DSm rule of combination and is not equivalent to Dempter's rule. It works for any models (the free DSm model, Shafer's model or any other hybrid models) when manipulating {\it{precise}} generalized (or eventually classical) basic belief functions. An extension of this rule for the combination of {\it{imprecise}} generalized (or eventually classical) basic belief functions is presented in next section. As already stated, in DSmT framework it is also possible to deal directly with complements if necessary depending on the problem under consideration and the information provided by the sources of evidence themselves. \\

The first and simplest way is to work with $S^\Theta$ on Shafer's model when a minimal refinement is possible and makes sense. The second way is to deal with partially known frame and introduce directly the complementary hypotheses into the frame itself. By example, if one knows only two hypotheses $\theta_1$, $\theta_2$ and their complements $\bar{\theta}_1$, $\bar{\theta}_2$, then we can choose switch from original frame $\Theta=\{\theta_1,\theta_2\}$ to the new frame $\Theta=\{\theta_1,\theta_2,\bar{\theta}_1,\bar{\theta}_2\}$. In such case, we don't necessarily assume that $\bar{\theta}_1=\theta_2$ and $\bar{\theta}_2=\theta_1$ because $\bar{\theta}_1$ and $\bar{\theta}_2$ may include other unknown hypotheses we have no information about (case of partial known frame). More generally, in DSmT framework, it is not necessary that the frame is built on pure/simple (possibly vague) hypotheses $\theta_i$ as usually done in all theories managing uncertainty. The frame $\Theta$ can also contain directly as elements conjunctions and/or disjunctions (or mixed propositions) and negations/complements of pure hypotheses as well. The DSm rules also work in such non-classic frames because DSmT works on any distributive lattice built from $\Theta$ anywhere $\Theta$ is defined.

\subsection{Examples of combination rules}

Here are some numerical examples on results obtained by DSm rules of combination. More examples can be found in \cite{Ref-DSmTBook_2004a}.

\subsubsection{Example with $\Theta=\{\theta_1,\theta_2,\theta_3,\theta_4\}$}

Let's consider the frame of discernment $\Theta=\{\theta_1,\theta_2,\theta_3,\theta_4\}$, two independent experts, and the two following bbas
$$m_1(\theta_1)=0.6\quad m_1(\theta_3)=0.4\quad m_2(\theta_2)=0.2\quad m_2(\theta_4)=0.8$$
\noindent represented in terms of mass matrix 
\begin{equation*}
\mathbf{M}=
\begin{bmatrix}
0.6 & 0  & 0.4 & 0\\
0 & 0.2 & 0 & 0.8
\end{bmatrix}
\end{equation*}
\begin{itemize}
\item Dempster's rule cannot be applied because: $\forall 1\leq j \leq 4$, one gets $m(\theta_j) = 0/0$ (undefined!).
\item But the classic DSm rule works because one obtains: $m(\theta_1) = m(\theta_2) = m(\theta_3) = m(\theta_4) = 0$, 
and $m(\theta_1 \cap \theta_2) = 0.12$, $m(\theta_1 \cap \theta_4) = 0.48$,
$m(\theta_2 \cap \theta_3) = 0.08$, $m(\theta_3 \cap \theta_4) = 0.32$ (partial paradoxes/conflicts).
\item Suppose now one finds out that all intersections are empty (Shafer's model), then one applies the hybrid DSm rule and one gets (index $h$ stands here for {\it{hybrid}} rule): $m_h(\theta_1\cup \theta_2)=0.12$, $m_h(\theta_1\cup\theta_4)=0.48$, $m_h(\theta_2\cup\theta_3)=0.08$ and $m_h(\theta_3\cup\theta_4)=0.32$.
\end{itemize}

\subsubsection{Generalization of Zadeh's example with $\Theta=\{\theta_1,\theta_2,\theta_3\}$}

Let's consider $0 < \epsilon_1,\epsilon_2 < 1$ be two very tiny positive numbers (close to zero), the frame of discernment be $\Theta=\{\theta_1,\theta_2,\theta_3\}$, have two experts (independent sources of evidence $s_1$ and $s_2$) giving the belief masses
$$m_1(\theta_1)=1-\epsilon_1 \quad m_1(\theta_2)=0 \quad m_1(\theta_3)=\epsilon_1$$
$$m_2(\theta_1)=0 \quad m_2(\theta_2)=1-\epsilon_2 \quad m_2(\theta_3)=\epsilon_2$$
\noindent From now on, we prefer to use matrices to describe the masses, i.e.
$$\begin{bmatrix}
1-\epsilon_1 & 0 &\epsilon_1\\
0 & 1-\epsilon_2 & \epsilon_2
\end{bmatrix}
$$
\begin{itemize}
\item Using Dempster's rule of combination, one gets
$$m(\theta_3)=\frac{(\epsilon_1\epsilon_2)}{(1-\epsilon_1)\cdot 0 + 0\cdot (1-\epsilon_2) + \epsilon_1\epsilon_2}=1$$
\noindent which is absurd (or at least counter-intuitive). Note that whatever positive values for $\epsilon_1$, $\epsilon_2$ are, Dempster's rule of combination provides always the same result (one) which is abnormal. The only acceptable and correct result obtained by Dempster's rule is really obtained only in the trivial case when $\epsilon_1=\epsilon_2=1$, i.e. when both sources agree in $\theta_3$ with certainty which is obvious.
\item Using the DSm rule of combination based on free-DSm model, one gets 
$m(\theta_3)=\epsilon_1\epsilon_2$, $m(\theta_1\cap \theta_2)=(1-\epsilon_1)(1-\epsilon_2)$, $m(\theta_1\cap \theta_3)=(1-\epsilon_1)\epsilon_2$, $m(\theta_2\cap \theta_3)=(1-\epsilon_2)\epsilon_1$ and the others are zero which appears more reliable/trustable.
\item Going back to Shafer's model and using the hybrid DSm rule of combination, one gets 
$m(\theta_3)=\epsilon_1\epsilon_2$, $m(\theta_1\cup \theta_2)=(1-\epsilon_1)(1-\epsilon_2)$, $m(\theta_1\cup \theta_3)=(1-\epsilon_1)\epsilon_2$, $m(\theta_2\cup \theta_3)=(1-\epsilon_2)\epsilon_1$ and the others are zero.
\end{itemize}

\noindent
Note that in the special  case when $\epsilon_1=\epsilon_2=1/2$, one has
$$m_1(\theta_1)=1/2 \quad m_1(\theta_2)=0 \quad m_1(\theta_3)=1/2$$
$$m_2(\theta_1)=0 \quad m_2(\theta_2)=1/2 \quad m_2(\theta_3)=1/2$$
Dempster's rule of combinations still yields $m(\theta_3)=1$ while the hybrid DSm rule based on the same Shafer's model yields now
$m(\theta_3)=1/4$, $m(\theta_1\cup \theta_2)=1/4$, $m(\theta_1\cup \theta_3)=1/4$, $m(\theta_2\cup \theta_3)=1/4$ which is normal.

\subsubsection{Comparison with Smets, Yager and Dubois \& Prade rules}
\label{Ref-Comparison}

We compare the results provided by DSmT rules and the main common rules of combination on the following very simple numerical example where only 2 independent sources (a priori assumed equally reliable) are involved and providing their belief initially on the 3D frame $\Theta=\{\theta_1,\theta_2,\theta_3\}$. It is assumed in this example that Shafer's model holds and thus the belief assignments $m_1(.)$ and $m_2(.)$ do not  commit belief to internal conflicting information. $m_1(.)$ and $m_2(.)$ are chosen as follows:
 $$m_1(\theta_1)=0.1 \qquad m_1(\theta_2)=0.4 \qquad m_1(\theta_3)=0.2 \qquad m_1(\theta_1\cup \theta_2)=0.3$$
 $$m_2(\theta_1)=0.5 \qquad m_2(\theta_2)=0.1 \qquad m_2(\theta_3)=0.3 \qquad m_2(\theta_1\cup \theta_2)=0.1$$
\noindent
These belief masses are usually represented in the form of a belief mass matrix $\mathbf{M}$ given by
\begin{equation}
\mathbf{M}=
\begin{bmatrix}
0.1 & 0.4 & 0.2 & 0.3\\
0.5 & 0.1 & 0.3 & 0.1
\end{bmatrix}
\end{equation}
\noindent
where index $i$ for the rows corresponds to the index of the source no. $i$ and the indexes $j$ for columns of $\mathbf{M}$ correspond to a given choice for enumerating the focal elements of all sources.
In this particular example, index $j=1$ corresponds to $\theta_1$, $j=2$ corresponds to $\theta_2$, $j=3$ corresponds to $\theta_3$ and  $j=4$ corresponds to $\theta_1\cup \theta_2$.\\

Now let's imagine that one finds out that $\theta_3$ is actually truly empty because some extra and certain knowledge on $\theta_3$ is received by the fusion center. As example, $\theta_1$, $\theta_2$ and $\theta_3$ may correspond to three suspects (potential murders) in a police investigation, $m_1(.)$ and $m_2(.)$ corresponds to two reports of independent witnesses, but it turns out that finally $\theta_3$ has provided a strong alibi to the criminal police investigator once arrested by the policemen. This situation corresponds to set up a hybrid model $\mathcal{M}$ with the constraint $\theta_3\overset{\mathcal{M}}{=}\emptyset$. \\

Let's examine the result of the fusion in such situation obtained by the Smets', Yager's, Dubois \& Prade's and hybrid DSm rules of combinations. First note that, based on the free DSm model, one would get by applying the classic DSm rule (denoted here by index $DSmC$) the following fusion result
\begin{align*}
m_{DSmC}(\theta_1)&=0.21 \qquad m_{DSmC}(\theta_2)=0.11 \\
m_{DSmC}(\theta_3)& =0.06 \qquad m_{DSmC}(\theta_1\cup\theta_2)=0.03\\
m_{DSmC}(\theta_1\cap\theta_2)&=0.21 \qquad m_{DSmC}(\theta_1\cap\theta_3)=0.13\\
m_{DSmC}(\theta_2\cap\theta_3)&=0.14 \qquad m_{DSmC}(\theta_3\cap(\theta_1\cup\theta_2))=0.11\\
\end{align*}

But because of the exclusivity constraints (imposed here by the use of Shafer's model and by the non-existential constraint $\theta_3\overset{\mathcal{M}}{=}\emptyset$), the total conflicting mass is actually given by $k_{12}=0.06 + 0.21 + 0.13 + 0.14 + 0.11=0.65$.
\begin{itemize}
\item If one applies  {\textbf{Dempster's rule}} \cite{Ref-Shafer_1976} (denoted here by index $DS$), one gets:
\begin{align*}
m_{DS}(\emptyset)& = 0\\
m_{DS}(\theta_1)&=0.21/[1- k_{12}]=0.21/[1-0.65]=0.21/0.35=0.600000\\
m_{DS}(\theta_2)&=0.11/[1-k_{12}]=0.11/[1-0.65]=0.11/0.35=0.314286\\
m_{DS}(\theta_1\cup\theta_2)&=0.03/[1-k_{12}]=0.03/[1-0.65]=0.03/0.35=0.085714
\end{align*}
\end{itemize}

\begin{itemize}
\item If one applies {\textbf{Smets' rule}} \cite{Ref-Smets_1994,Ref-Smets_2000} (i.e. the non normalized version of Dempster's rule with the conflicting mass transferred onto the empty set), one gets:
\begin{align*}
m_{S}(\emptyset)&=m(\emptyset)=0.65 \qquad\text{(conflicting mass)}\\
m_{S}(\theta_1)&=0.21\\
m_{S}(\theta_2)&=0.11\\
m_{S}(\theta_1\cup\theta_2)&=0.03
\end{align*}
\end{itemize}

\begin{itemize}
\item If one applies {\textbf{Yager's rule}} \cite{Ref-Yager_1983,Ref-Yager_1985,Ref-Yager_1987}, one gets:
\begin{align*}
m_{Y}(\emptyset)&= 0\\
m_{Y}(\theta_1)&=0.21\\
m_{Y}(\theta_2)&=0.11\\
m_{Y}(\theta_1\cup\theta_2)&=0.03 + k_{12}=0.03+0.65=0.68
\end{align*}
\end{itemize}

\begin{itemize}
\item If one applies {\textbf{Dubois \& Prade's rule}} \cite{Ref-Dubois_1988}, one gets because $\theta_3\overset{\mathcal{M}}{=}\emptyset$ :
\begin{align*}
m_{DP}(\emptyset)& = 0 \qquad \text{(by definition of Dubois \& Prade's rule)}\\
m_{DP}(\theta_1)&= [m_1(\theta_1)m_2(\theta_1) + m_1(\theta_1)m_2(\theta_1\cup\theta_2)\\
& \quad + m_2(\theta_1)m_1(\theta_1\cup\theta_2)]\\
& \quad + [m_1(\theta_1)m_2(\theta_3) + m_2(\theta_1)m_1(\theta_3)]\\
& = [0.1\cdot 0.5+0.1\cdot 0.1 +0.5\cdot 0.3] + [0.1\cdot 0.3 +0.5\cdot 0.2] \\
& = 0.21 + 0.13 = 0.34\\
m_{DP}(\theta_2)&=[0.4\cdot 0.1 + 0.4\cdot 0.1 +0.1\cdot 0.3] + [0.4\cdot 0.3 + 0.1\cdot 0.2] \\
& = 0.11 + 0.14 = 0.25
\end{align*}
\begin{align*}
m_{DP}(\theta_1\cup\theta_2)&=[m_1(\theta_1\cup\theta_2)m_2(\theta_1\cup\theta_2)] \\
& \quad + [m_1(\theta_1\cup\theta_2)m_2(\theta_3) + m_2(\theta_1\cup\theta_2)m_1(\theta_3)]\\
& \quad + [m_1(\theta_1)m_2(\theta_2) + m_2(\theta_1)m_1(\theta_2)]\\
& = [0.3 0.1 ] + [0.3  \cdot 0.3 + 0.1  \cdot 0.2 ] + [0.1 \cdot 0.1 + 0.5  \cdot 0.4] \\
& = [0.03] + [0.09+0.02] + [0.01 + 0.20]\\
& = 0.03 + 0.11 + 0.21 = 0.35
\end{align*}
Now if one adds up the masses, one gets $0+ 0.34+0.25+0.35=0.94$ which is less than 1. Therefore Dubois \& Prade's rule of combination does not work when a singleton, or an union of singletons, becomes empty (in a dynamic fusion problem). The products of such empty-element columns of the mass matrix $\mathbf{M}$ are lost; this problem is fixed in DSmT by the sum $S_2(.)$ in \eqref{Chapter1eq:DSmHkBis1} which transfers these products to the total or partial ignorances.
\end{itemize}

\begin{itemize}
\item Finally, if one applies {\textbf{DSmH rule}}, one gets because $\theta_3\overset{\mathcal{M}}{=}\emptyset$ :
\begin{align*}
m_{DSmH}(\emptyset)& = 0 \qquad \text{(by definition of DSmH)}\\
m_{DSmH}(\theta_1)&= 0.34 \qquad \text{(same as $m_{DP}(\theta_1)$)}\\
m_{DSmH}(\theta_2)&= 0.25 \qquad \text{(same as $m_{DP}(\theta_2)$)}\\
m_{DSmH}(\theta_1\cup\theta_2)&=[m_1(\theta_1\cup\theta_2)m_2(\theta_1\cup\theta_2)] \\
& \quad + [m_1(\theta_1\cup\theta_2)m_2(\theta_3) + m_2(\theta_1\cup\theta_2)m_1(\theta_3)]\\
& \quad + [m_1(\theta_1)m_2(\theta_2) + m_2(\theta_1)m_1(\theta_2)] + [m_1(\theta_3)m_2(\theta_3)] \\
& = 0.03 + 0.11 + 0.21 + 0.06 = 0.35+0.06 = 0.41\\
& \neq m_{DP}(\theta_1\cup\theta_2)
\end{align*}
We can easily verify that $m_{DSmH}(\theta_1) + m_{DSmH}(\theta_2)+ m_{DSmH}(\theta_1\cup\theta_2)=1$. In this example, using the hybrid DSm rule, one transfers the product of the empty-element $\theta_3$ column, $m_1(\theta_3)m_2(\theta_3)=0.2\cdot 0.3=0.06$, to $m_{DSmH}(\theta_1\cup\theta_2)$, which becomes equal to $0.35+0.06=0.41$.
Clearly, DSmH rule doesn't provide the same result as Dubois and Prade's rule, but only when working on static frames of discernment (restricted cases).
\end{itemize}

\subsection{Fusion of imprecise beliefs}
\label{Chapter1sec2.7}

In many fusion problems, it seems very difficult (if not impossible) to have precise sources of evidence generating precise basic belief assignments (especially when belief functions are provided by human experts), and a more flexible plausible and paradoxical  theory supporting imprecise information becomes necessary. In the previous sections, we presented the fusion of {\it{precise}} uncertain and conflicting/paradoxical generalized basic belief assignments (gbba) in DSmT framework. We mean here by precise gbba, basic belief functions/masses $m(.)$ defined precisely on the hyper-power set  $D^\Theta$ where each mass $m(X)$, where $X$ belongs to $D^\Theta$, is represented by only one real number belonging to $[0,1]$ such that $\sum_{X\in D^\Theta}m(X)=1$. In this section, we present the DSm fusion rule for dealing with {\it{admissible imprecise generalized basic belief assignments}} $m^I(.)$ defined as real subunitary intervals of $[0,1]$, or even more general as real subunitary sets [i.e. 
sets, not necessarily intervals]. \\

An imprecise belief assignment $m^I(.)$ over $D^\Theta$ is said {\textit{admissible}} if and only if there exists for every $X\in D^\Theta$ at least one real number $m(X)\in m^I(X)$ such that $\sum_{X\in D^\Theta}m(X)=1$. The idea to work with imprecise belief structures represented by real subset intervals of $[0,1]$ is not new and has been investigated in \cite{Ref-Lamata_1994,Ref-Denoeux_1997,Ref-Denoeux_1999} and references therein. The proposed works available in the literature, upon our knowledge were limited only to sub-unitary interval combination in the framework of Transferable Belief Model (TBM) developed by Smets \cite{Ref-Smets_1994,Ref-Smets_2000}. We extend the approach of Lamata \& Moral and Den\oe ux based on subunitary interval-valued masses to subunitary set-valued masses; therefore the 
closed intervals used by Den\oe ux to denote imprecise masses are generalized to any sets included in [0,1], i.e. in our case these sets can be unions of (closed, open, or half-open/half-closed) intervals and/or scalars all in $[0,1]$. Here, the proposed extension is done in the context of DSmT framework, although it can also apply directly to fusion of imprecise belief structures within TBM as well if the user prefers to adopt TBM rather than DSmT.\\

Before presenting the general formula for the combination of generalized imprecise belief structures, we remind the following set operators involved in the DSm fusion formulas. Several numerical examples are given in the chapter 6 of \cite{Ref-DSmTBook_2004a}.

\begin{itemize}
 \item
 {\textbf{Addition of sets}}
 \begin{equation*}
 S_{1}\boxplus S_{2} =S_{2}\boxplus S_{1}\triangleq \{ x \mid x = s_{1}+s_{2},  s_{1} \in 
S_{1},s_{2} \in S_{2} \}
\label{Chapter1eq:addition}
\end{equation*}
\item
{\textbf{Subtraction of sets}}
 \begin{equation*}
 S_{1}\boxminus S_{2} \triangleq \{ x \mid x = s_{1}-s_{2},  s_{1} \in 
S_{1}, s_{2} \in S_{2} \}
 \label{Chapter1eq:addition}
 \end{equation*}
  \item
 {\textbf{Multiplication of sets}}
 \begin{equation*}
S_{1}\boxdot S_{2} \triangleq \{ x \mid x = s_{1}\cdot s_{2}, s_{1} \in 
S_{1},s_{2} \in S_{2} \}
\label{Chapter1eq:multiplication}
 \end{equation*}
\end{itemize}
 
\begin{itemize}
 \item {\textbf{Division of sets}}: If $0$ doesn't belong to $S_2$,
  \begin{equation*}
S_{1}\boxslash S_{2} \triangleq \{ x \mid x = s_{1}/s_{2}, 
 s_{1} \in S_{1}, s_{2} \in S_{2}\}\label{Chapter1eq:divisionofsets}
 \end{equation*}
 
 \end{itemize}

\subsubsection{DSm rule of combination for imprecise beliefs}
We present the generalization of the DSm rules to combine any type of imprecise belief assignment which may be represented by the union of several sub-unitary (half-) open intervals, (half-)closed intervals and/or sets of points belonging to [0,1]. Several numerical examples are also given. In the sequel, one uses the notation $(a,b)$ for an open interval, $[a,b]$ for a closed interval, and $(a,b]$ or $[a,b)$ for a half open and half closed interval. From the previous operators on sets, one can generalize the DSm rules (classic and hybrid) from scalars to sets in the following way \cite{Ref-DSmTBook_2004a} (chap. 6): $\forall A\neq\emptyset \in D^\Theta$,
\begin{equation}
m^I(A) = \underset{\underset{(X_1\cap X_2\cap\ldots\cap X_k)=A}{X_1,X_2,\ldots,X_k\in D^\Theta}}{\boxed{\sum}}
\underset{i=1,\ldots,k}{\boxed{\prod}} m_i^I(X_i)
\label{Chapter1eq:DSMruleSetsImprecise}
\end{equation}
\noindent
where $\boxed{\sum}$ and $\boxed{\prod}$ represent the summation, and respectively product, of sets.\\

Similarly, one can generalize the hybrid DSm rule from scalars to sets in the following way:
\begin{equation}
m^I_{DSmH}(A) = m_{\mathcal{M}(\Theta)}^I(A)\triangleq 
\phi(A)\boxdot \Bigl[ S_1^I(A) \boxplus S_2^I(A) \boxplus S_3^I(A)\Bigr]
 \label{Chapter1eq:DSmHkBisImprecise}
\end{equation}
\noindent
where all sets involved in formulas are in the canonical form and $\phi(A)$ is the {\it{characteristic non emptiness function}} of the set $A$ and $S_1^I(A)$, $S_2^I(A)$ and $S_3^I(A)$ are defined by
\begin{equation}
S_1^I(A)\triangleq
\underset{\underset{X_1\cap X_2\cap\ldots\cap X_k=A}{X_1,X_2,\ldots,X_k\in D^\Theta}}{\boxed{\sum}}
\underset{i=1,\ldots,k}{\boxed{\prod}} m_i^I(X_i)
\label{Chapter1eq:S1I}
\end{equation}
\begin{equation}
S_2^I(A)\triangleq 
\underset{\underset{[\mathcal{U}=A]\vee [(\mathcal{U}\in\boldsymbol{\emptyset}) \wedge (A=I_t)]}{X_1,X_2,\ldots,X_k\in\boldsymbol{\emptyset}}}{\boxed{\sum}}
\underset{i=1,\ldots,k}{\boxed{\prod}} m_i^I(X_i)
\label{Chapter1eq:S2I}
\end{equation}
\begin{equation}
S_3^I(A)\triangleq
\underset{\underset{X_1\cap X_2\cap\ldots\cap X_k\in\boldsymbol{\emptyset} }{\underset{X_1\cup X_2\cup\ldots\cup X_k=A}{X_1,X_2,\ldots,X_k\in D^\Theta}}}{\boxed{\sum}}
\underset{i=1,\ldots,k}{\boxed{\prod}} m_i^I(X_i)
\label{Chapter1eq:S3I}
\end{equation}
In the case when all sets are reduced to points (numbers), the set operations become normal operations with numbers; the sets operations are generalizations of numerical operations. When imprecise belief structures reduce to precise belief structure, DSm rules  \eqref{Chapter1eq:DSMruleSetsImprecise} and \eqref{Chapter1eq:DSmHkBisImprecise} reduce to their precise version  \eqref{Chapter1JDZT}  and  \eqref{Chapter1eq:DSmHkBis1} respectively.

\subsubsection{Example}

Here is a simple example of fusion with multiple-interval masses. For simplicity, this example is a particular case when the theorem of admissibility (see \cite{Ref-DSmTBook_2004a} p. 138 for details) is verified by a few points, which happen to be just on the bounders. It is an extreme example, because we tried to comprise all kinds of possibilities which may occur in the imprecise or very imprecise fusion. So, let's consider a fusion problem over $\Theta=\{\theta_1,\theta_2\}$, two independent sources of information with the following imprecise admissible belief assignments
\begin{table}[h]
\begin{equation*}
\begin{array}{|c|c|c|}
\hline
A\in D^\Theta & m_1^I(A) & m_2^I(A) \\
\hline
\theta_1 & [0.1,0.2] \cup \{0.3\} & [0.4,0.5]\\
\theta_2 &(0.4,0.6)\cup [0.7,0.8] &  [0,0.4]\cup \{0.5,0.6\}\\
\hline
\end{array}
\end{equation*}
\caption{Inputs of the fusion with imprecise bba's.}
\label{Chapter1mytablex1}
\end{table}

\noindent
Using the DSm classic (DSmC) rule for sets, one gets
\begin{align*}
m^I(\theta_1) & = ([0.1,0.2] \cup \{0.3\})\boxdot [0.4,0.5] \\
&= ([0.1,0.2] \boxdot [0.4,0.5])\cup (\{0.3\}\boxdot [0.4,0.5] )\\
& = [0.04,0.10] \cup [0.12,0.15]
\end{align*}
\begin{align*}
m^I(\theta_2)& =((0.4,0.6)\cup [0.7,0.8] )\boxdot ([0,0.4]\cup \{0.5,0.6\})\\
&= ((0.4,0.6)\boxdot [0,0.4])\cup ((0.4,0.6)\boxdot  \{0.5,0.6\})\\
&\qquad \cup ([0.7,0.8]\boxdot [0,0.4]) \cup ([0.7,0.8]\boxdot \{0.5,0.6\})\\
&= (0,0.24)\cup (0.20,0.30) \cup (0.24,0.36)\cup [0,0.32] \\
& \qquad \cup [0.35,0.40] \cup [0.42,0.48] = [0,0.40] \cup [0.42,0.48]
\end{align*}
\begin{align*}
m^I(\theta_1\cap \theta_2)&=
[([0.1,0.2] \cup \{0.3\})\boxdot([0,0.4]\cup \{0.5,0.6\})] \boxplus [[0.4,0.5]\\
& \quad \boxdot ((0.4,0.6)\cup [0.7,0.8]) ]\\
&=[ ([0.1,0.2]\boxdot [0,0.4]) \cup ([0.1,0.2]\boxdot \{0.5,0.6\}) \\
& \quad \cup ( \{0.3\}\boxdot [0,0.4]) \cup ( \{0.3\}\boxdot  \{0.5,0.6\})] \\
& \quad \boxplus [ ([0.4,0.5]\boxdot (0.4,0.6)) \cup ([0.4,0.5]\boxdot [0.7,0.8] ) ] \\
& = [[0,0.08]\cup [0.05,0.10]\cup [0.06,0.12] \cup [0,0.12] \\
& \quad \cup \{0.15,0.18\}] \boxplus [(0.16,0.30)\cup[0.28,0.40]]\\
&= [[0,0.12]\cup \{0.15,0.18\}]\boxplus (0.16,0.40] \\
& =(0.16,0.52] \cup (0.31,0.55] \cup (0.34,0.58]=(0.16,0.58]
\end{align*}

\noindent
Hence finally the fusion admissible result with DSmC rule is given by:
\begin{table}[!h]
\begin{equation*}
\begin{array}{|c|c|}
\hline
A\in D^\Theta & m^I(A)= [m_1^I \oplus m_2^I](A) \\
\hline
\theta_1 & [0.04,0.10] \cup [0.12,0.15]\\
\theta_2 & [0,0.40] \cup [0.42,0.48] \\
\theta_1\cap \theta_2  & (0.16,0.58]\\
\theta_1\cup \theta_2 & 0 \\
\hline
\end{array}
\end{equation*}
\caption{Fusion result with the DSmC rule.}
\label{Chapter1mytablex2}
\end{table}

\noindent
If one finds out\footnote{We consider now a dynamic fusion problem.} that $\theta_1\cap \theta_2 \overset{\mathcal{M}}{\equiv}\emptyset$ (this is our hybrid model $\mathcal{M}$ one wants to deal with), then one uses the hybrid DSm rule for sets \eqref{Chapter1eq:DSmHkBisImprecise}: $m_{\mathcal{M}}^I(\theta_1\cap \theta_2)=0$ and $m_{\mathcal{M}}^I(\theta_1\cup \theta_2)= (0.16,0.58]$, the others imprecise masses are not changed.

\clearpage
\newpage
With the hybrid DSm rule (DSmH) applied to imprecise beliefs, one gets now the results given in Table \ref{Chapter1mytablex3}.

\begin{table}[!h]
\begin{equation*}
\begin{array}{|c|c|}
\hline
A\in D^\Theta & m_{\mathcal{M}}^I(A)= [m_1^I \oplus m_2^I](A) \\
\hline
\theta_1 & [0.04,0.10] \cup [0.12,0.15]\\
\theta_2 & [0,0.40] \cup [0.42,0.48] \\
\theta_1\cap \theta_2\overset{\mathcal{M}}{\equiv}\emptyset  & 0 \\
\theta_1\cup \theta_2 & (0.16,0.58]\\
\hline
\end{array}
\end{equation*}
\caption{Fusion result with DSmH rule for $\mathcal{M}$.}
\label{Chapter1mytablex3}
\end{table}

Let's check now the admissibility condition. For the source 1, there exist the precise masses $(m_1(\theta_1)=0.3) \in ([0.1,0.2] \cup \{0.3\})$ and $(m_1(\theta_2)=0.7) \in ((0.4,0.6)\cup [0.7,0.8])$ such that $0.3+0.7=1$. For the source 2, there exist the precise masses $(m_1(\theta_1)=0.4) \in ([0.4,0.5])$ and $(m_2(\theta_2)=0.6) \in ([0,0.4]\cup \{0.5,0.6\})$ such that $0.4+0.6=1$. Therefore both sources associated with $m_1^I(.)$ and $m_2^I(.)$ are admissible imprecise sources of information. It can be verified that DSmC fusion of $m_1(.)$ and $m_2(.)$ yields the paradoxical bba $m(\theta_1)=[m_1\oplus m_2](\theta_1)=0.12$, $m(\theta_2)=[m_1\oplus m_2](\theta_2)=0.42$ and $m(\theta_1\cap \theta_2)=[m_1\oplus m_2](\theta_1\cap \theta_2)=0.46$. One sees that the admissibility condition is satisfied since $(m(\theta_1)=0.12)\in (m^I(\theta_1)=[0.04,0.10] \cup [0.12,0.15])$, $(m(\theta_2)=0.42)\in (m^I(\theta_2)=[0,0.40] \cup [0.42,0.48])$ and $(m(\theta_1\cap \theta_2)=0.46)\in (m^I(\theta_1\cap \theta_2)=(0.16,0.58])$ such that $0.12+0.42+0.46=1$. Similarly if one finds out that $\theta_1\cap\theta_2=\emptyset$, then one uses DSmH rule and one gets: $m(\theta_1\cap\theta_2)=0$ and $m(\theta_1\cup\theta_2)=0.46$; the others remain unchanged. The admissibility condition still holds, because one can pick at least one number in each subset $m^I(.)$ such that the sum of these numbers is  1. 

\section{Proportional Conflict Redistribution rule}

Instead of applying a direct transfer of partial conflicts onto partial uncertainties as with DSmH, the idea behind the Proportional Conflict Redistribution (PCR) rule \cite{Ref-Smarandache_2005c,Ref-Book_2006} is to transfer (total or partial) conflicting masses to non-empty sets involved in the conflicts proportionally with respect to the masses assigned to them by sources as follows:

\begin{enumerate}
\item calculation the conjunctive rule of the belief masses of sources;
\item calculation the total or partial conflicting masses;
\item redistribution of the (total or partial) conflicting masses to the non-empty sets involved in the conflicts proportionally with respect to their masses assigned by the sources.
\end{enumerate}
The way the conflicting mass is redistributed yields actually several versions of PCR rules. These PCR fusion rules work for any degree of conflict, for any DSm models (Shafer's model, free DSm model or any hybrid DSm model) and both in DST and DSmT frameworks for static or dynamical fusion situations. We present below only the most sophisticated proportional conflict redistribution rule denoted PCR5 in \cite{Ref-Smarandache_2005c,Ref-Book_2006}. PCR5 rule is what we feel the most efficient PCR fusion rule developed so far. This rule redistributes the partial conflicting mass to the elements involved in the partial conflict, considering the conjunctive normal form of the partial conflict. PCR5 is what we think the most mathematically exact redistribution of conflicting mass to non-empty sets following the logic of the conjunctive rule. It does a better redistribution of the conflicting mass than Dempster's rule since PCR5 goes backwards on the tracks of the conjunctive rule and redistributes the conflicting mass only to the sets involved in the conflict and proportionally to their masses put in the conflict. PCR5 rule is quasi-associative and preserves the neutral impact of the vacuous belief assignment because in any partial conflict, as well in the total conflict (which is a sum of all partial conflicts), the conjunctive normal form of each partial conflict does not include $\Theta$ since $\Theta$ is a neutral element for intersection (conflict), therefore $\Theta$ gets no mass after the redistribution of the conflicting mass. We have proved in \cite{Ref-Book_2006} the continuity property of the fusion result with continuous variations of bba's to combine.  

\subsection{PCR formulas}

The PCR5 formula for the combination of two sources ($s=2$) is given by: $m_{PCR5}(\emptyset)=0$ and $\forall X\in {G^\Theta}\setminus\{\emptyset\}$
\begin{equation}
m_{PCR5}(X)=m_{12}(X) +
\sum_{\substack{Y\in {G^\Theta}\setminus\{X\} \\ X\cap Y=\emptyset}} 
[\frac{m_1(X)^2m_2(Y)}{m_1(X)+m_2(Y)} +
 \frac{m_2(X)^2 m_1(Y)}{m_2(X)+m_1(Y)}]
   \label{Chapter1eq:PCR5}
 \end{equation}
 
\medskip
\noindent
where all sets involved in formulas are in canonical form and where $G^\Theta$ corresponds to classical power set $2^\Theta$ if Shafer's model is used, or to a constrained hyper-power set $D^\Theta$ if any other hybrid DSm model is used instead, or to the super-power set $S^\Theta$ if the minimal refinement $\Theta^{ref}$ of $\Theta$ is used; $m_{12}(X)\equiv m_{\cap}(X)$ corresponds to the conjunctive consensus on $X$ between the $s=2$ sources and where all denominators are different from zero. If a denominator is zero, that fraction is discarded.\\

A general formula of PCR5 for the fusion of $s>2$ sources has been proposed in \cite{Ref-Book_2006}, but a more intuitive PCR formula (denoted PCR6) which provides good results in practice has been proposed by Martin and Osswald in \cite{Ref-Book_2006} (pages 69-88) and is given by: $m_{PCR6}(\emptyset)=0$ and $\forall X\in {G^\Theta}\setminus\{\emptyset\}$

\begin{equation}
\label{PaperIntroDSmT-PCR6}
 m_{PCR6}(X)  =  \displaystyle m_{12\ldots s}(X) + 
  \sum_{i=1}^s m_i(X)^2 
  \sum_{
  \begin{array}{c}
      \scriptstyle {\displaystyle \mathop{\cap}_{k=1}^{s\!-\!1}} Y_{\sigma_i(k)} \cap X \equiv \emptyset \\
      \scriptstyle (Y_{\sigma_i(1)},...,Y_{\sigma_i(s\!-\!1)})\in (G^\Theta)^{s\!-\!1}
  \end{array}
  }
  \left(\!\!\frac{\displaystyle \prod_{j=1}^{s\!-\!1} m_{\sigma_i(j)}(Y_{\sigma_i(j)})}
       {\displaystyle m_i(X) \!+\! \sum_{j=1}^{s\!-\!1}
  m_{\sigma_i(j)}(Y_{\sigma_i(j)})}\!\!\right)
\end{equation}

\noindent
where $\sigma_i$ counts from 1 to $s$ avoiding $i$:
\begin{eqnarray}
\label{MOt:sigma}
\left\{
\begin{array}{ll}
\sigma_i(j)=j &\mbox{if~} j<i,\\
\sigma_i(j)=j+1 &\mbox{if~} j\geq i,\\
\end{array}
\right.
\end{eqnarray}

\noindent
Since $Y_i$ is a focal element of expert/source $i$, $\scriptstyle m_i(X)+\displaystyle \sum_{j=1}^{s-1} m_{\sigma_i(j)}(Y_{\sigma_i(j)}) \neq 0$; the belief
mass assignment $m_{12\ldots s}(X)\equiv m_{\cap}(X)$ corresponds to the conjunctive consensus on $X$ between the $s>2$ sources.
For two sources ($s=2$), PCR5 and PCR6 formulas coincide.

\subsection{Examples}

\begin{itemize}
\item {\textbf{Example 1}}: Let's take $\Theta=\{A, B\}$ of exclusive elements (Shafer's model), and the following bba:
\begin{center}
\begin{tabular}[h]{|c|ccc|}
\hline
 & $A$ & $B$ & $ A\cup B$\\
 \hline
 $m_1(.)$ & 0.6 & 0 & 0.4 \\
 \hline
 $m_2(.)$ & 0 & 0.3 & 0.7 \\
 \hline
 \hline
 $m_{\cap}(.)$ & 0.42 & 0.12 & 0.28 \\
 \hline
\end{tabular}
\end{center}
The conflicting mass is $k_{12}=m_{\cap}(A\cap B)$ and equals $m_1(A)m_2(B)+m_1(B)m_2(A)=0.18$.
Therefore $A$ and $B$ are the only focal elements involved in the conflict. Hence according to the PCR5 hypothesis only $A$ and $B$ deserve a part of the conflicting mass and $A\cup B$ do not deserve. With PCR5, one redistributes the conflicting mass $k_{12}=0.18$ to $A$ and $B$ proportionally with the masses $m_1(A)$
and $m_2(B)$ assigned to $A$ and $B$ respectively. 
\clearpage
\newpage

\noindent
Here are the results obtained from Dempster's rule, DSmH and PCR5:
\begin{center}
\begin{tabular}[h]{|l||ccc|}
\hline
  & $A$ & $B$ & $A\cup B$\\
 \hline
  $m_{DS}$ & 0.512 & 0.146 & 0.342 \\
  $m_{DSmH}$ & 0.420 & 0.120 & 0.460 \\
  $m_{PCR5}$ & 0.540 & 0.180 &  0.280 \\
\hline
\end{tabular}
\end{center}
\noindent

\item {\textbf{Example 2}}: Let's modify example 1 and consider
\begin{center}
\begin{tabular}[h]{|c|ccc|}
\hline
 & $A$ & $B$ & $ A\cup B$\\
 \hline
 $m_1(.)$ & 0.6 & 0 & 0.4 \\
 \hline
 $m_2(.)$ & 0.2 & 0.3 & 0.5 \\
 \hline
 \hline
 $m_{\cap}(.)$ & 0.50 & 0.12 & 0.20 \\
 \hline
\end{tabular}
\end{center}
\noindent The conflicting mass $k_{12}=m_{\cap}(A\cap B)$ as well as the distribution coefficients for the PCR5 remains the same as in the previous example but one gets now
\begin{center}
\begin{tabular}[h]{|l||ccc|}
\hline
  & $A$ & $B$ & $A\cup B$\\
 \hline
  $m_{DS}$ & 0.609 & 0.146 & 0.231 \\
 $m_{DSmH}$ & 0.500 & 0.120 & 0.380 \\
 $m_{PCR5}$ & 0.620 & 0.180 &  0.200 \\
\hline
\end{tabular}
\end{center}

\item {\textbf{Example 3}}: Let's modify example 2 and consider
\begin{center}
\begin{tabular}[h]{|c|ccc|}
\hline
 & $A$ & $B$ & $ A\cup B$\\
 \hline
 $m_1(.)$ & 0.6 & 0.3 & 0.1 \\
 \hline
 $m_2(.)$ & 0.2 & 0.3 & 0.5 \\
 \hline
 \hline
 $m_{\cap}(.)$ & 0.44 & 0.27 & 0.05 \\
 \hline
\end{tabular}
\end{center}
The conflicting mass $k_{12}=0.24 = m_1(A)m_2(B)+m_1(B)m_2(A)=0.24$ is now different from previous examples, which means that $m_2(A) = 0.2$ and $m_1(B) =0.3$ did make an impact on the conflict. Therefore $A$ and $B$ are the only focal elements involved in the conflict and thus only $A$ and $B$ deserve a part of the conflicting mass. PCR5 redistributes the partial conflicting mass 0.18 to $A$ and $B$ proportionally with the masses $m_1(A)$ and $m_2(B)$ and also the partial conflicting mass 0.06 to $A$ and $B$ proportionally with the masses $m_2(A)$ and $m_1(B)$. After all derivations (see \cite{Ref-Florea2007} for details), one finally gets:
\begin{center}
\begin{tabular}[h]{|l||ccc|}
\hline
  & $A$ & $B$ & $A\cup B$\\
 \hline
  $m_{DS}$ & 0.579 & 0.355 & 0.066 \\
 $m_{DSmH}$ & 0.440 & 0.270 & 0.290 \\
 $m_{PCR5}$ & 0.584 & 0.366 &  0.050 \\
\hline
\end{tabular}
\end{center}
\noindent

One clearly sees that $m_{DS}(A\cup B)$ gets some mass from the conflicting mass although $A\cup B$ does not deserve any part of the conflicting mass (according to PCR5 hypothesis) since $A\cup B$ is not involved in the conflict (only $A$ and $B$ are involved in the conflicting mass). Dempster's rule appears to us less exact than PCR5 and Inagaki's rules \cite{Ref-Inagaki_1991}. It can be showed  \cite{Ref-Florea2007} that Inagaki's fusion rule (with an optimal choice of tuning parameters) can become in some cases very close to PCR5 but upon our opinion PCR5 result is more exact (at least less ad-hoc than Inagaki's one).
\end{itemize}

\begin{itemize}
\item {\textbf{Example 4 (A more concrete example)}}: Three people, John ($J$), George ($G$), and David ($D$) are suspects to a murder.  So the frame of discernment is  $\Theta \triangleq \{J, G, D\}$. Two sources $m_{1}(.)$ and $m_{2}(.)$ (witnesses) provide the following information:

\begin{center}
\begin{tabular}[h]{|l||ccc|}
\hline
  & $J$ & $G$ & $D$\\
 \hline
  $m_{1}$ & 0.9 & 0 & 0.1 \\
 $m_{2}$ & 0 & 0.8 & 0.2 \\
\hline
\end{tabular}
\end{center}
\noindent
%

We know that John and George are friends, but John and David hate each other, and similarly George and David.
\begin{itemize}
\item[a)] Free model, i. e. all intersections are nonempty: $J\cap G \neq \emptyset$, $J\cap D \neq \emptyset$, $G\cap D \neq \emptyset$, 
$J\cap G\cap D \neq \emptyset$. Using the DSm classic rule one gets:

\begin{center}
\begin{tabular}[h]{|l||ccccccc|}
\hline
  & $J$ & $G$ & $D$ & $J\cap G$ & $J\cap D$ & $G\cap D$ & $J \cap G\cap D$\\
 \hline
  $m_{DSmC}$ & 0 & 0 & 0.02 & 0.72 & 0.18 & 0.08 & 0 \\
 \hline
\end{tabular}
\end{center}


So we can see that John and George together ($J\cap G$) are most likely to have committed the crime, since the mass $m_{DSmC}(J\cap G)=0.72$ is the biggest resulting mass after the fusion of the two sources.  In Shafer's model, only one suspect could commit the crime, but the free and hybrid models allow two or more people to have committed the same crime - which happens in reality.

\item[b)] Let's consider the hybrid model, i. e. some intersections are empty, and others are not. According to the above statement about the relationships between the three suspects, we can deduce that $J\cap G\neq \emptyset$, while $J\cap D = G\cap D = J\cap G\cap D = \emptyset$.
Then we first apply the DSm Classic rule, and then the transfer of the conflicting masses is done with PCR5:

\begin{center}
\small
\begin{tabular}[h]{|l||ccccccc|}
\hline
  & $J$ & $G$ & $D$ & $J\cap G$ & $J\cap D$ & $G\cap D$ & $J \cap G\cap D$\\
 \hline
  $m_{1}$ & 0.9 & 0 & 0.1 &  &  &  &  \\
 $m_{2}$ & 0 & 0.8 & 0.2 &  &  &  &  \\
  \hline
  $m_{DSmC}$ & 0 & 0 & 0.02 & 0.72 & 0.18 & 0.08 & 0 \\
 \hline
\end{tabular}
\end{center}

Using PCR5 now we transfer $m(J\cap D)=0.18$, since $J\cap D=\emptyset$, to $J$ and $D$ proportionally with 0.9 and 0.2 respectively,
so $J$ gets 0.15 and $D$ gets 0.03 since:
$$xJ/0.9 = z1D/0.2 = 0.18/(0.9+0.2) = 0.18/1.1$$

\noindent
whence $xJ = 0.9(0.18/1.1) = 0.15$ and $z1D = 0.2(0.18/1.1) = 0.03$.

Again using PCR5, we transfer $m(G\cap D)=0.08$,  since $G\cap D=\emptyset$, to $G$ and $D$ proportionally with 0.8 and 0.1 respectively, so $G$ gets 0.07 and $D$ gets 0.01 since:

$$yG/0.8 = z2D/0.1 = 0.08/(0.8+0.1) = 0.08/0.9$$

\noindent whence $yG = 0.8(0.08/0.9) = 0.07$ and $zD = 0.1(0.08/0.9) = 0.01$. Adding we get finally:

\begin{center}
\small
\begin{tabular}[h]{|l||ccccccc|}
\hline
  & $J$ & $G$ & $D$ & $J\cap G$ & $J\cap D$ & $G\cap D$ & $J \cap G\cap D$\\
 \hline
  $m_{PCR5}$ & 0.15 & 0.07 & 0.06 & 0.72 & 0 & 0 & 0 \\
 \hline
\end{tabular}
\end{center}

So one has a high belief that the criminals are John and George (both of them committed the crime) since $m(J\cap D) = 0.72$ and it is by far the greatest fusion mass.
\end{itemize}

In Shafer’s model, if we try to refine we get the disjoint parts: $D$, $J\cap G$, $J\setminus (J\cap G)$, and $G\setminus (J\cap G)$, but the last two are ridiculous (what is the real/physical nature of $J\setminus (J\cap G)$ or $G\setminus (J\cap G)$ ? Half of a person(!) ?), so the refining does not work here in reality. That’s why the hybrid and free models are needed.

\end{itemize}

\begin{itemize}
\item {\textbf{Example 5 (Imprecise PCR5)}}: The PCR5 formula can naturally work also for the combination of imprecise bba's. This has been already presented in section 1.11.8 page 49 of \cite{Ref-Book_2006} with a numerical example to show how to apply it. This example will therefore not be reincluded here.
\end{itemize}

\clearpage
\newpage

\subsection{Zadeh's example}

We compare here the solutions for well-known Zadeh's example \cite{Ref-Zadeh_1979,Ref-Zadeh_1986} provided by several fusion rules. A detailed presentation with more comparisons can be found in \cite{Ref-DSmTBook_2004a,Ref-Book_2006}. Let's consider $\Theta=\{M,C,T\}$ as the frame of three potential origins about possible diseases of a patient ($M$ standing for {\it{meningitis}}, $C$ for {\it{concussion}} and $T$ for {\it{tumor}}), the Shafer's model and the two following belief assignments provided by two independent doctors after examination of the same patient.
\begin{align*}
m_1(M)&=0.9 &\quad m_1(C)&=0 &\quad m_1(T)&=0.1\\
m_2(M)&=0 &\quad m_2(C)&=0.9 &\quad m_2(T)&=0.1
\end{align*}
The total conflicting mass is high since it is
$$m_1(M)m_2(C)+m_1(M)m_2(T)+m_2(C)m_1(T)=0.99$$
\begin{itemize}
\item with Dempster's rule and Shafer's model (DS), one gets the counter-intuitive result (see justifications in \cite{Ref-Zadeh_1979,Ref-Dubois_1986c,Ref-Yager_1987,Ref-Voorbraak_1991,Ref-DSmTBook_2004a}): $m_{DS}(T)=1$
\item with Yager's rule \cite{Ref-Yager_1987} and Shafer's model: 
$m_{Y}(M\cup C \cup T)=0.99$ and $m_{Y}(T)=0.01$
\item with DSmH and Shafer's model:
$$m_{DSmH}(M\cup C)=0.81\qquad  m_{DSmH}(T)=0.01$$
$$m_{DSmH}(M\cup T)=m_{DSmH}(C\cup T)=0.09$$
\item The Dubois \& Prade's rule (DP) \cite{Ref-Dubois_1986c} based on Shafer's model provides in Zadeh's example the same result as DSmH, because DP and DSmH coincide in all static fusion problems\footnote{Indeed DP rule has been developed for static fusion only while DSmH has been developed to take into account the possible dynamicity of the frame itself and also its associated model.}.
\item with PCR5 and Shafer's model: $m_{PCR5}(M)=m_{PCR5}(C)=0.486$ and $m_{PCR5}(T)=0.028$.
\end{itemize}
One sees that when the total conflict between sources becomes high, DSmT is able (upon authors opinion) to manage more adequately through DSmH or PCR5 rules the combination of information than Dempster's rule, even when working with Shafer's model - which is only a specific hybrid model. DSmH rule is in agreement with DP rule for the static fusion, but DSmH and DP rules differ in general (for non degenerate cases) for dynamic fusion while PCR5 rule is the most exact proportional conflict redistribution rule. Besides this particular example, we showed in \cite{Ref-DSmTBook_2004a} that there exist several infinite classes of counter-examples to Dempster's rule which can be solved by DSmT. \medskip

In summary, DST based on Dempster's rule  provides counter-intuitive results in Zadeh's example, or in non-Bayesian examples similar to Zadeh's and no result when the conflict is 1. Only ad-hoc discounting techniques allow to circumvent troubles of Dempster's rule or we need to switch to another model of representation/frame; in the later case the solution obtained doesn't fit with the Shafer's model one originally wanted to work with. We want also to emphasize that in dynamic fusion when the conflict becomes high, both DST \cite{Ref-Shafer_1976} and Smets' Transferable Belief Model (TBM) \cite{Ref-Smets_1994} approaches fail to respond to new information provided by new sources. This can be easily showed by the very simple following example. \medskip

\noindent
{\textbf{Example}} (where TBM doesn't respond to new information): \\

\noindent
Let $\Theta=\{A,B,C\}$ with the (precise) bba's $m_1(A)=0.4$, $m_1(C)=0.6$ and $m_2(A)=0.7$, $m_2(B)=0.3$. Then one gets\footnote{We introduce here explicitly the indexes of sources in the fusion result since more than two sources are considered in this example.} with Dempster's rule, Smets' TBM (i.e. the non-normalized version of Dempster's combination), DSmH and PCR5: $m_{DS}^{12}(A)=1$, $m_{TBM}^{12}(A)=0.28$, $m_{TBM}^{12}(\emptyset)=0.72$, 
\begin{equation*}
\begin{cases}
m_{DSmH}^{12}(A)=0.28\\
m_{DSmH}^{12}(A\cup B)=0.12\\
m_{DSmH}^{12}(A\cup C)=0.42\\
m_{DSmH}^{12}(B\cup C)=0.18
\end{cases}
\qquad \text{and} \qquad
\begin{cases}
m_{PCR5}^{12}(A)=0.574725\\
m_{PCR5}^{12}(B)=0.111429\\
m_{PCR5}^{12}(C)=0.313846
\end{cases}
\end{equation*}

Now let's consider a temporal fusion problem and introduce a third source $m_3(.)$ with $m_3(B)=0.8$ and $m_3(C)=0.2$. Then one sequentially combines the results obtained by $m_{TBM}^{12}(.)$, $m_{DS}^{12}(.)$, $m_{DSmH}^{12}(.)$ and $m_{PCR}^{12}(.)$ with
the new evidence $m_3(.)$ and one sees that $m_{DS}^{(12)3}$ becomes not defined (division by zero) and $m_{TBM}^{(12)3}(\emptyset)=1$ while (DSmH) and (PCR5) provide 
\begin{equation*}
\begin{cases}
m_{DSmH}^{(12)3}(B)=0.240\\
m_{DSmH}^{(12)3}(C)=0.120\\
m_{DSmH}^{(12)3}(A\cup B)=0.224\\
m_{DSmH}^{(12)3}(A\cup C)= 0.056\\
m_{DSmH}^{(12)3}(A\cup B\cup C)= 0.360\\
\end{cases}
\qquad \text{and} \qquad
\begin{cases}
m_{PCR5}^{(12)3}(A)=0.277490\\
m_{PCR5}^{(12)3}(B)=0.545010\\
m_{PCR5}^{(12)3}(C)=0.177500
\end{cases}
\end{equation*}

When the mass committed to empty set becomes one at a previous temporal fusion step, then both DST and TBM do not respond to new information. Let's continue the example and consider a fourth source $m_4(.)$ with $m_4(A)=0.5$, $m_4(B)=0.3$ and $m_4(C)=0.2$. Then it is easy to see that $m_{DS}^{((12)3)4}(.)$ is not defined since at previous step $m_{DS}^{(12)3}(.)$ was already not defined, and that $m_{TBM}^{((12)3)4}(\emptyset)=1$ whatever $m_4(.)$ is because at the previous fusion step one had $m_{TBM}^{(12)3}(\emptyset)=1$. Therefore for a number of sources $n\geq 2$, DST and TBM approaches do not respond to new information incoming in the fusion process while both (DSmH) and (PCR5) rules respond to new information. To make DST and/or TBM working properly in such cases, it is necessary to introduce ad-hoc temporal discounting techniques which are not necessary to introduce if DSmT is adopted. If there are good reasons to introduce temporal discounting, there is obviously no difficulty to apply the DSm fusion of these discounted sources. An analysis of this behavior for target type tracking is presented in \cite{Ref-Dezert_2006,Ref-Book_2006}.

\section{The generalized pignistic transformation (GPT)}
\label{PaperIntroDSmT-GPT section}

\subsection{The classical pignistic transformation}

We follow here Philippe Smets'  vision which considers the management of information as a two 2-levels process: credal (for combination of evidences) and pignistic\footnote{Pignistic terminology has been coined by Philippe Smets and comes from {\it{pignus}}, a bet in Latin.} (for decision-making) , i.e "{\it{when someone must take a decision, he/she must then construct a probability function derived from the belief function that describes his/her credal state. This probability function is then used to make decisions}}" \cite{Ref-Smets_1988} (p. 284). One obvious way to build this probability function corresponds to the so-called Classical Pignistic Transformation (CPT) defined in DST framework (i.e. based on the Shafer's model assumption) as \cite{Ref-Smets_2000}:

\begin{equation}
BetP\{A\}=\sum_{X \in 2^\Theta}\frac{|X\cap A|}{|X|}m(X)
\label{Chapter1eq:Pig}
\end{equation}

\noindent
where $|A|$ denotes the number of worlds in the set $A$ (with convention $|\emptyset | / |\emptyset |=1$, to define $BetP\{\emptyset \}$). Decisions are achieved by computing the expected utilities of the acts using the subjective/pignistic $BetP\{.\}$ as the probability function needed to compute expectations.
Usually, one uses the maximum of the pignistic probability as decision criterion. The maximum of $BetP\{.\}$ is often considered as a prudent betting decision criterion between the two other alternatives (max of plausibility or max. of credibility which appears to be respectively too optimistic or too pessimistic). It is easy to show that $BetP\{.\}$ is indeed a probability function (see \cite{Ref-Smets_1994}).

\subsection{Notion of DSm cardinality}
One important notion involved in the definition of the Generalized Pignistic Transformation (GPT) is the {\it{DSm cardinality}}. The {\it{DSm cardinality}} of any element $A$ of hyper-power set  $D^\Theta$, denoted $\mathcal{C}_\mathcal{M}(A)$, corresponds to the number of parts of $A$ in the corresponding fuzzy/vague Venn diagram of the problem (model $\mathcal{M}$) taking into account the set of integrity constraints (if any), i.e. all the possible intersections due to the nature of the elements $\theta_i$. This {\it{intrinsic cardinality}} depends on the model $\mathcal{M}$ (free, hybrid or Shafer's model).  $\mathcal{M}$ is the model that contains $A$, which depends both on the
dimension $n=\vert \Theta \vert$ and on the number of non-empty intersections present in its associated Venn diagram (see \cite{Ref-DSmTBook_2004a} for details ). The DSm cardinality depends on the cardinal of $\Theta = \{\theta_1,\theta_2,\ldots,\theta_n\}$ and on the model of $D^\Theta$ (i.e., the number of intersections and between what elements of $\Theta$ - in a word the structure) at the same time; it is not necessarily that every singleton, say $\theta_i$, has the same DSm cardinal, because each singleton has a different structure; if its structure is the simplest (no intersection of this elements with other elements) then $\mathcal{C}_\mathcal{M}(\theta_i)=1$, if the structure is more complicated (many intersections) then $\mathcal{C}_\mathcal{M}(\theta_i) > 1$; let's consider a singleton $\theta_i$: if it has 1 intersection only then $\mathcal{C}_\mathcal{M}(\theta_i)=2$, for 2 intersections only $\mathcal{C}_\mathcal{M}(\theta_i)$ is 3 or 4 depending on the model $\mathcal{M}$, for $m$ intersections it is between $m+1$ and $2^m$ depending on the model; the maximum DSm cardinality is $2^{n-1}$ and occurs for $\theta_1\cup\theta_2\cup\ldots\cup\theta_n$ in the free model $\mathcal{M}^f$; similarly for any set from $D^\Theta$: the more complicated structure it has, the bigger is the DSm cardinal;
thus the DSm cardinality measures the complexity of en element from $D^\Theta$, which is a nice characterization in our opinion; we may say that for the singleton $\theta_i$ not even $\vert\Theta\vert$ counts, but only its structure (= how many other singletons intersect $\theta_i$). Simple illustrative examples are given in Chapter 3 and 7 of \cite{Ref-DSmTBook_2004a}.
One has $1 \leq \mathcal{C}_\mathcal{M}(A) \leq 2^n-1$. $\mathcal{C}_\mathcal{M}(A)$ must not be confused with the classical cardinality $\vert A \vert$ of a given set $A$ (i.e. the number of its distinct elements) - that's why a new notation is necessary here.
$\mathcal{C}_\mathcal{M}(A)$ is very easy to compute by programming from the algorithm of generation of $D^\Theta$ given explicated in \cite{Ref-DSmTBook_2004a}.\\

\noindent
{\textbf{Example}}: let's take back the example of the simple hybrid DSm model described in section \ref{Chapter1Sec:DSMmodels}, then one gets the following list of elements (with their DSm cardinal) for the restricted $D^\Theta$ taking into account the integrity constraints of this hybrid model:
\begin{equation*}
\begin{array}{lcl}
A\in D^\Theta                     & \mathcal{C}_{\mathcal{M}}(A) \\
\hline
\alpha_0\triangleq\emptyset                                                 & 0 \\
\alpha_1\triangleq\theta_1\cap\theta_2                             & 1 \\
\alpha_2\triangleq\theta_3                                                    & 1 \\
\alpha_3\triangleq\theta_1                                                    & 2 \\
\alpha_4\triangleq\theta_2                                                    & 2 \\
\alpha_5\triangleq\theta_1\cup\theta_2                             & 3 \\
\alpha_6\triangleq\theta_1\cup\theta_3                             & 3 \\
\alpha_7\triangleq\theta_2\cup\theta_3                             & 3 \\
\alpha_8\triangleq\theta_1\cup\theta_2\cup\theta_3       & 4  \\
\end{array}
\end{equation*}
\begin{center}
{\textit{Example of DSm cardinals}}: $\mathcal{C}_{\mathcal{M}}(A)$ for hybrid model $\mathcal{M}$.
\end{center}

\subsection{The  Generalized Pignistic Transformation}

To take a rational decision within DSmT framework, it is necessary to generalize the Classical Pignistic Transformation in order to construct a pignistic probability function from any generalized basic belief assignment $m(.)$ drawn from the DSm rules of combination. Here is the simplest and direct extension of the CPT to define the Generalized Pignistic Transformation:
\begin{equation}
\forall A \in D^\Theta, \qquad \qquad BetP\{A\}=\sum_{X \in D^\Theta}  \frac{\mathcal{C}_{\mathcal{M}}(X\cap A)}{\mathcal{C}_{\mathcal{M}}(X)}m(X)
\label{Chapter1eq:PigG}
\end{equation}
\noindent
where $\mathcal{C}_{\mathcal{M}}(X)$ denotes the DSm cardinal of proposition $X$ for the DSm model $\mathcal{M}$ of the problem under consideration.\\

The decision about the solution of the problem  is usually taken by the maximum of pignistic probability function $BetP\{.\}$. Let's remark the close ressemblance of the two pignistic transformations \eqref{Chapter1eq:Pig} and \eqref{Chapter1eq:PigG}. It can be shown that \eqref{Chapter1eq:PigG} reduces to \eqref{Chapter1eq:Pig} when the hyper-power set $D^\Theta$ reduces to classical power set $2^\Theta$ if we adopt Shafer's model. But  \eqref{Chapter1eq:PigG} is a generalization of  \eqref{Chapter1eq:Pig} since it can be used for computing pignistic probabilities for any models (including Shafer's model). It has been proved in \cite{Ref-DSmTBook_2004a} (Chap. 7) that $BetP\{.\}$ defined in \eqref{Chapter1eq:PigG} is indeed a probability distribution. In the following section, we introduce a new alternative to BetP which is presented in details in \cite{Ref-Book_2009}.

\section{The DSmP transformation}

In the theories of belief functions, the mapping from the belief to the probability domain is a controversial issue. The original purpose of such mappings was to make (hard) decision, but contrariwise to erroneous widespread idea/claim,  this is not the only interest for using such mappings nowadays. Actually the probabilistic transformations of belief mass assignments (as the pignistic transformation mentioned previously) are for example very useful in modern multitarget multisensor tracking systems (or in any other systems) where one deals with soft decisions (i.e. where all  possible solutions are kept for state estimation with their likelihoods). For example, in a Multiple Hypotheses Tracker using both kinematical and attribute data, one needs to compute all probabilities values for deriving the likelihoods of data association hypotheses and then mixing them altogether to estimate states of targets. Therefore, it is very relevant to use a mapping which provides a high probabilistic information content (PIC) for expecting better performances. \\

In this section, we briefly recall a new probabilistic transformation, denoted $DSmP$ and introduced in \cite{Ref-DezertSmarandacheDSmP2008} which will be explained in details in \cite{Ref-Book_2009}. $DSmP$ is straight and different from other transformations. The basic idea of $DSmP$ consists in a new way of proportionalizations of the mass of  each partial ignorance such as $A_1\cup A_2$ or $A_1\cup (A_2\cap A_3)$  or $(A_1\cap A_2)\cup (A_3\cap A_4)$, etc. and the mass of the total ignorance $A_1\cup A_2\cup \ldots \cup A_n$, to the elements involved in the ignorances. This new transformation takes into account both the values of the masses and the cardinality of elements in the proportional redistribution process. We first remind what PIC criteria is and then shortly present the general formula for DSmP transformation with few numerical examples. More examples and comparisons with respect to other transformations are given in \cite{Ref-Book_2009}.

 \subsection{The Probabilistic Information Content (PIC)}

Following Sudano's approach \cite{Ref-Sudano2001,Ref-Sudano2002,Ref-Sudano2006}, we adopt the Probabilistic Information Content (PIC) criterion as a metric depicting the strength of a critical decision by a specific probability distribution. It is an essential measure in any threshold-driven automated decision system. The PIC is the dual of the normalized Shannon entropy. A PIC value of one indicates the total knowledge to make a correct decision (one hypothesis has a probability value of one and the rest of zero). A PIC value of zero indicates that the knowledge to make a correct decision does not exist (all the hypotheses have an equal probability value), i.e. one has the maximal entropy. The PIC is used in our analysis to sort the performances of the different pignistic transformations through several numerical examples. We first recall what Shannon entropy and PIC measure are and their tight relationship.

\begin{itemize}
\item \textbf{Shannon entropy}
\end{itemize}

Shannon entropy, usually expressed in bits (binary digits), of a probability measure $P\{.\}$ over a discrete finite set $\Theta=\{\theta_1,\ldots,\theta_n\}$ is defined by\footnote{with common convention $0\log_2 0 = 0$.} \cite{Ref-Shannon48}:
\begin{equation}
\label{Chapter1ShannonEntropy}
H(P)\triangleq - \sum_{i=1}^{n} P\{\theta_i\} \log_2(P\{\theta_i\})
\end{equation}
$H(P)$ is maximal for the uniform probability distribution over $\Theta$, i.e. when $P\{\theta_i\}=1/n$ for $i=1,2,\ldots, n$. In that case, one gets $H(P)=H_{\max} =  - \sum_{i=1}^{n} \frac{1}{n} \log_2(\frac{1}{n})= \log_2(n)$. $H(P)$ is minimal for a totally {\it{deterministic}} probability, i.e. for any $P\{.\}$ such that $P\{\theta_i\}=1$ for some $i\in\{1,2,\ldots,n\}$ and $P\{\theta_j\}=0$ for $j\neq i$. $H(P)$ measures the randomness carried by any discrete probability $P\{.\}$. 

\begin{itemize}
\item \textbf{The PIC metric}
\end{itemize}
\label{Chapter1SectionPIC}
The Probabilistic Information Content (PIC) of a probability measure $P\{.\}$ associated with a probabilistic source over a discrete finite set $\Theta=\{\theta_1,\ldots,\theta_n\}$ is defined by \cite{Ref-Sudano2002}:
\begin{equation}
\label{Chapter1PIC}
PIC(P)=1 + \frac{1}{H_{\max}} \cdot \sum_{i=1}^{n} P\{\theta_i\} \log_2(P\{\theta_i\})
\end{equation}
The PIC is nothing but the dual of the normalized Shannon entropy and thus is actually unit less. $PIC(P)$ takes its values in $[0,1]$. $PIC(P)$ is maximum, i.e. $PIC_{\max}=1$ with any {\it{deterministic}} probability and it is minimum, i.e. $PIC_{\min}=0$, with the uniform probability over the frame $\Theta$. The simple relationships between $H(P)$ and $PIC(P)$ are $PIC(P)=1 - (H(P)/H_{\max})$ and $H(P)=H_{\max}\cdot (1-PIC(P))$.

 \subsection{ The DSmP formula}

Let's consider a discrete frame $\Theta$ with a given model (free DSm model, hybrid DSm model or Shafer's model), the $DSmP$ mapping is defined by
$DSmP_{\epsilon}(\emptyset)=0$ and $\forall X \in G^\Theta \setminus \{\emptyset\}$ by
\begin{equation}
DSmP_{\epsilon}(X)=\sum_{Y \in G^\Theta}  \frac{\displaystyle\sum_{\substack{Z \subseteq X\cap Y \\ \mathcal{C}(Z)=1}} m(Z) + \epsilon\cdot \mathcal{C}(X\cap Y)}{\displaystyle\sum_{\substack{Z \subseteq Y \\ \mathcal{C}(Z)=1}} m(Z) + \epsilon\cdot \mathcal{C}(Y)}m(Y)
\label{Chapter1eq:DSmP}
\end{equation}
\noindent
where $\epsilon\geq 0$ is a tuning parameter and $G^\Theta$ corresponds to the generic set ($2^\Theta$, $S^\Theta$ or $D^\Theta$ including eventually all the integrity constraints (if any) of the model $\mathcal{M}$); $\mathcal{C}(X\cap Y)$ and $\mathcal{C}(Y)$ denote the DSm cardinals\footnote{We have omitted the index of the model $\mathcal{M}$ for the notation convenience.} of the sets $X\cap Y$ and $Y$ respectively. $\epsilon$ allows to reach the maximum PIC value of the approximation of $m(.)$ into a subjective probability measure. The smaller $\epsilon$, the better/bigger PIC value. In some particular degenerate cases however, the $DSmP_{\epsilon=0}$ values cannot be derived, but the $DSmP_{\epsilon>0}$ values can however always be derived by choosing $\epsilon$ as a very small positive number, say $\epsilon=1/1000$ for example in order to be as close as we want to the maximum of the PIC. 
When $\epsilon=1$ and when the masses of all elements $Z$ having $\mathcal{C}(Z)=1$ are zero, \eqref{Chapter1eq:DSmP} reduces to \eqref{Chapter1eq:PigG}, i.e. $DSmP_{\epsilon=1}= BetP$. The passage from a free DSm model to a Shafer's model
involves the passage from a structure to another one, and the cardinals change as well in the formula \eqref{Chapter1eq:DSmP}. \\

$DSmP$ works for all models (free, hybrid and Shafer's). In order to apply classical transformation (Pignistic, Cuzzolin's one, Sudano's ones, etc - see \cite{Ref-Book_2009}), we need at first to refine the frame (on the cases when it is possible!) in order to work with Shafer's model, and then apply their formulas. In the case where refinement makes sense, then one can apply the other subjective probabilities on the refined frame.  $DSmP$ works on the refined frame as well and gives the same result as it does on the non-refined frame. Thus $DSmP$ with $\epsilon>0$ works on any models and so is very general and appealing. $DSmP$ does a redistribution of the ignorance mass with respect to both the singleton masses and the singletons' cardinals in the same time. Now, if all masses of singletons involved in all ignorances are different from zero, then we can take $\epsilon=0$, and $DSmP$ gives the best result, i.e. the best PIC value. 
In summary, $DSmP$ does an 'improvement' over previous known probabilistic transformations in the sense that $DSmP$ mathematically makes a more accurate redistribution of the ignorance masses to the singletons involved in ignorances. $DSmP$ and $BetP$ work in both theories: DST (= Shafer's model) and DSmT (= free or hybrid models) as well.

\clearpage
\newpage

 \subsection{Examples for DSmP and BetP}
\label{Chapter1IntroChapterExample4}
 
The examples briefly presented here are detailed in \cite{Ref-Book_2009}  which includes also additional results based on Cuzzolin's and Sudano's transformations.
 
 \begin{itemize}
 \item \textbf{With Shafer's model and a non-Bayesian mass}
 \end{itemize}
 
Let's consider the frame $\Theta=\{A,B\}$ and let's assume Shafer's model and the non-Bayesian mass (more precisely the simple support mass) given in Table \ref{Chapter1MassExample4}. We summarize in Table \ref{Chapter1TableExample4}, the results obtained with DSmP and BetP. One sees that $PIC(DSmP_{\epsilon\rightarrow 0})$ is maximum among all PIC values. 

 \begin{table}[!h]
\centering
 \begin{tabular}{|l|c|c|c|}
    \hline
              & $A$ & $B$ & $A\cup B$ \\
    \hline
$m(.)$  & 0.4 & 0 & 0.6\\
    \hline
  \end{tabular}
  \caption{Quantitative inputs.}
\label{Chapter1MassExample4}
\end{table}

\begin{table}[!h]
\centering
\begin{tabular}{|l|c|c||c|}
    \hline
              & $A$ & $B$  & $PIC(.)$ \\
    \hline
   $BetP(.)$     & 0.7000 & 0.3000 & 0.1187 \\
  $DSmP_{\epsilon=0.001}(.)$  & 0.9985 & 0.0015 & 0.9838 \\
  $DSmP_{\epsilon=0}(.)$  & 1 & 0 & 1 \\
   \hline
  \end{tabular}
   \caption{Results of the probabilistic transformations.}
\label{Chapter1TableExample4}
\end{table}

The best result is an {\it{adequate probability}}, not {\it{the biggest PIC}} in this case.
This is because $P(B)$ deserves to receive some mass from $m(A\cup B)$, so the most correct result is done by $DSmP_{\epsilon=0.001}$ in Table \ref{Chapter1TableExample4} (of course we can choose any other very small positive value for $\epsilon$ if we want). Always when a singleton whose mass is zero, but it is involved in an ignorance whose mass is not zero, then $\epsilon$ (in $DSmP$ formula \eqref{Chapter1eq:DSmP}) should be taken different from zero.

 \begin{itemize}
 \item \textbf{With a Hybrid DSm model}
 \end{itemize}
\label{Chapter1IntroChapterExample8}

\begin{figure}[!h]
\begin{center}
{\tt \setlength{\unitlength}{1pt}
\begin{picture}(90,90)
\thinlines    
\put(40,60){\circle{40}}
\put(60,60){\circle{40}}
\put(50,10){\circle{40}}
\put(15,84){\vector(1,-1){10}}
\put(7,84){$A$}
\put(84,84){\vector(-1,-1){10}}
\put(85,84){$B$}
\put(85,10){\vector(-1,0){15}}
\put(87,7){$C$}
\put(45,59){$D'$}
\put(47,7){$C'$}
\put(65,59){$B'$}
\put(28,59){$A'$}
\end{picture}}
\end{center}
\caption{Hybrid model for $\Theta=\{A,B,C\}$.}
\label{Chapter1FigHybridDSmModelRefined}
\end{figure}
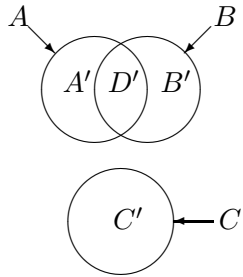

Let's consider the frame $\Theta=\{A,B,C\}$ and let's consider the hybrid DSm model in which all intersections of elements of $\Theta$ are empty, but $A\cap B$ corresponding to figure \ref{Chapter1FigHybridDSmModelRefined}. In this case, $G^\Theta$ reduces to 9 elements $\{\emptyset, A\cap B, A, B, C, A\cup B, A\cup C, B\cup C, A\cup B \cup C\}$. The input masses of focal elements are given by $m(A\cap B)=0.20$, $m(A)=0.10$, $m(C)=0.20$, $m(A\cup B)=0.30$, $m(A\cup C)=0.10$, and $m(A\cup B\cup C)=0.10$ and given in the Table \ref{Chapter1MassExample8refined}. 

\begin{table}[!h]
\centering
\centering
 \begin{tabular}{|l|c|c|c|}
    \hline
     & $D'$ & $A'\cup D'$  & $C'$\\
    \hline
$m(.)$  & 0.2 & 0.1 & 0.2 \\
    \hline
   \hline
     & $A'\cup B'\cup D'$ & $A'\cup C'\cup D'$  & $A'\cup B'\cup C'\cup D'$ \\
    \hline
  $m(.)$  & 0.3 &  0.1 & 0.1\\
    \hline
  \end{tabular}
  \caption{Quantitative inputs.}
\label{Chapter1MassExample8refined}
\end{table}

\clearpage
\newpage

Applying BetP and DSmP transformations, one gets:

 \begin{table}[!h]
\centering
\begin{tabular}{|l|c|c|c|c||c|}
    \hline
                            & $A'$ & $B'$  & $C'$ & $D'$ & $PIC(.)$ \\
    \hline
       $BetP(.)$     &  0.2084 & 0.1250 & 0.2583 &  0.4083 & 0.0607 \\
       $DSmP_{\epsilon =0.001}(.)$      & 0.0025 & 0.0017 &  0.2996 &  0.6962 & 0.5390\\
   \hline
  \end{tabular}
   \caption{Results of the probabilistic transformations.}
\label{Chapter1TableExample8}
\end{table}
\vspace{-0.7cm}

 \begin{itemize}
 \item \textbf{With a free DSm model}
 \end{itemize}
\label{Chapter1IntroChapterExample9}

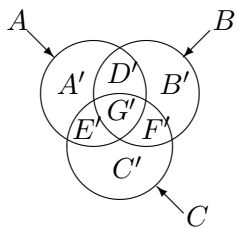
\begin{figure}[!h]
\begin{center}
{\tt \setlength{\unitlength}{1pt}
\begin{picture}(90,90)
\thinlines    
\put(40,60){\circle{40}}
\put(60,60){\circle{40}}
\put(50,40){\circle{40}}
\put(15,84){\vector(1,-1){10}}
\put(7,84){$A$}
\put(84,84){\vector(-1,-1){10}}
\put(85,84){$B$}
\put(74,15){\vector(-1,1){10}}
\put(75,10){$C$}
\put(45,64){$D'$}
\put(45,50){$G'$}
\put(47,28){$C'$}
\put(32,43){$E'$}
\put(58,43){$F'$}
\put(65,60){$B'$}
\put(26,60){$A'$}
\end{picture}}
\end{center}
\vspace{-0.5cm}
\caption{Free DSm model for a 3D frame.}
\label{Chapter1FigFree3DDSmModel}
\end{figure}

Let's consider the frame $\Theta=\{A,B,C\}$ and let's consider the free DSm model depicted on Figure \ref{Chapter1FigFree3DDSmModel} with the input masses given in Table \ref{Chapter1MassExample9}. To apply Sudano's and Cuzzolin's mappings, one works on the refined frame $\Theta^\text{ref}=\{A',B',C',D',E',F',G'\}$ where the elements of $\Theta^\text{ref}$ are exclusive (assuming such refinement has a physically sense) according to Figure \ref{Chapter1FigFree3DDSmModel}. This refinement step is not necessary when using $DSmP$ since it works directly on DSm free model. The PIC values obtained with DSmP and BetP are given in Table \ref{Chapter1TableExample9}. One sees that $DSmP_{\epsilon \rightarrow 0}$ provides here again the best results in term of PIC. 
\vspace{-0.5cm}
\begin{table}[!h]
\centering
 \begin{tabular}{|l|c|c|c|}
    \hline
     & $A\cap B \cap C$ & $A\cap B$  & $A$\\
    \hline
$m(.)$  & 0.1 & 0.2 & 0.3 \\
    \hline
   \hline
     & $A\cup B$ & $A\cup B \cup C$  &  \\
    \hline
  $m(.)$  & 0.1 &  0.3 &  \\
    \hline
  \end{tabular}
  \caption{Quantitative inputs.}
\label{Chapter1MassExample9}
\end{table}
\vspace{-0.5cm}
 \begin{table}[!h]
\centering
\begin{tabular}{|l|c|}
    \hline
        Transformations                    & $PIC(.)$ \\
    \hline
       $BetP(.)$     &  0.1176\\
       $DSmP_{\epsilon =0.001}(.)$      & 0.8986\\
   \hline
  \end{tabular}
   \caption{Results of the probabilistic transformations.}
\label{Chapter1TableExample9}
\end{table}

An extension of DSmP (denoted qDSmP) for working with qualitative labels instead of numbers is possible and has been proposed and presented in 2008 in \cite{Ref-DezertSmarandacheDSmP2008}. A simple example for qDSmP is given in the next section.

\section{Fusion of qualitative beliefs}

We recall here the notion of qualitative belief assignment to model beliefs of human experts expressed in natural language (with linguistic labels). We show how qualitative beliefs can be efficiently combined using an extension of DSmT to qualitative reasoning. A more detailed presentation can be found in \cite{Ref-Book_2006}. The derivations are based on a new arithmetic on linguistic labels which allows a direct extension of all quantitative rules of combination and conditioning. The qualitative version of PCR5 rule and DSmP is also presented in the sequel.
 
\subsection{Qualitative Operators}

Computing with words (CW) and qualitative information is more vague, less precise than computing with numbers, but it offers the advantage of robustness if done correctly. Here is a general arithmetic we propose for computing with words (i.e. with linguistic labels). Let's consider a finite frame $\Theta=\{\theta_1,\ldots,\theta_n\}$ of $n$ (exhaustive) elements $\theta_i$, $i=1,2,\ldots,n$, with an associated model $\mathcal{M}(\Theta)$ on $\Theta$ (either Shafer's model $\mathcal{M}^0(\Theta)$, free-DSm model $\mathcal{M}^f(\Theta)$, or more general any Hybrid-DSm model \cite{Ref-DSmTBook_2004a}). A model $\mathcal{M}(\Theta)$ is defined by the set of integrity constraints on elements of $\Theta$ (if any); Shafer's model $\mathcal{M}^0(\Theta)$ assumes all elements of $\Theta$ truly exclusive, while free-DSm model $\mathcal{M}^f(\Theta)$ assumes no exclusivity constraints between elements of the frame $\Theta$. Let's define a finite set of linguistic labels\index{linguistic labels} $\tilde{L}=\{L_1,L_2,\ldots,L_m\}$ where $m\geq 2$ is an integer. $\tilde{L}$ is endowed with a total order relationship $\prec$, so that $L_1\prec L_2\prec \ldots\prec L_m$. To work on a close linguistic set under linguistic addition and multiplication operators, we extends $\tilde{L}$ with two extreme values $L_{0}$ and $L_{m+1}$ where $L_{0}$ corresponds to the minimal qualitative value and $L_{m+1}$ corresponds to the maximal qualitative value, in such a way that
$$L_0\prec L_1\prec L_2\prec \ldots\prec L_m\prec L_{m+1}$$
\noindent
where $\prec$ means inferior to, or less (in quality) than, or smaller (in quality) than, etc. hence a relation of order from a qualitative point of view. But if we make a correspondence between qualitative
labels and quantitative values on the scale $[0, 1]$, then $L_{\min}=L_0$ would correspond to the numerical value 0, while $L_{\max}=L_{m+1}$ would correspond to the
numerical value 1, and each $L_i$ would belong to $[0,1]$, i. e.
$$L_{\min}=L_0 < L_1 < L_2 < \ldots <L_m < L_{m+1}=L_{\max}$$

\noindent
From now on, we work on extended ordered set $L$ of qualitative values
$$L=\{L_0,\tilde{L},L_{m+1}\}=\{L_0,L_1,L_2,\ldots,L_m,L_{m+1}\}$$
 
In our previous works, we did propose approximate qualitative operators, but in \cite{Ref-Book_2009} we propose to use better and accurate operators for qualitative labels.
Since these new operators are  defined in details in the chapter of \cite{Ref-Book_2009} devoted on the DSm Field and Linear Algebra of Refined Labels (FLARL), we just briefly introduce here only the  the main ones (i.e. the accurate label addition, multiplication and division). In FLARL, we can replace the "qualitative quasi-normalization" of qualitative operators we used in our previous papers by "qualitative normalization" since in FLARL we have exact qualitative calculations and exact normalization.

\begin{itemize}
\item Label addition :
\begin{equation}
L_{a}+L_{b} = L_{a+b}
\label{Ref-eq-LabelAddition}
\end{equation}
\noindent
since $\frac{a}{m+1}+\frac{b}{m+1}=\frac{a+b}{m+1}$.
\label{Chapter1qaddaccurate}
\item Label multiplication :
\begin{equation}
L_{a}\times L_{b} = L_{(a b)/(m+1)}
\label{Ref-eq-LabelMultiplication}
\end{equation}
\noindent
since $\frac{a}{m+1}\cdot \frac{b}{m+1}=\frac{(a b)/(m+1)}{m+1}$.

\item Label division (when $L_b\neq L_0$):
\begin{equation}
L_{a}\div L_{b} = L_{(a / b)(m+1)}
\label{Ref-eq-LabelDivision}
\end{equation}
\noindent
since $\frac{a}{m+1} \div \frac{b}{m+1}=\frac{a}{b} = \frac{(a / b) (m+1)}{m+1}$.
\end{itemize}

More accurate qualitative operations (substraction, scalar multiplication, scalar root, scalar power, etc) can be found in \cite{Ref-Book_2009}. Of course, if one really need to stay within the original set of labels, an approximation will be necessary at the very end of the calculations.

\subsection{Qualitative Belief Assignment}
A qualitative belief assignment\index{qualitative belief assignment}\footnote{We call it also {\it{qualitative belief mass}} or {\it{q-mass}} for short.} (qba) is a mapping function $qm(.): G^\Theta \mapsto L$ where $G^\Theta$ corresponds either to $2^\Theta$, to $D^\Theta$ or even to $S^\Theta$ depending on the model of the frame $\Theta$ we choose to work with. In the case when the labels are equidistant, i.e. the qualitative distance between any two consecutive labels is the same, we get an exact qualitative result, and a qualitative basic belief assignment (bba) is considered normalized if the sum of all its qualitative masses is equal to $L_{\max}=L_{m+1}$. If the labels are not equidistant, we still can use all qualitative operators defined in the FLARL, but the qualitative result is approximate, and a qualitative bba is considered quasi-normalized if the sum of all its masses is equal to $L_{\max}$. Using the qualitative operator of FLARL, we can easily extend all the combination and conditioning rules from quantitative to qualitative. In the sequel we will consider $s\geq 2$ qualitative belief assignments\index{qualitative belief assignment} $qm_1(.),\ldots, qm_s(.)$ defined over the same space $G^\Theta$ and provided by $s$ independent sources $S_1,\ldots,S_s$ of evidence. \\

 \noindent{\textbf{Note}}: The addition and multiplication operators used in all qualitative fusion formulas in next sections correspond to {\it{qualitative addition}} and {\it{qualitative multiplication}} operators and must not be confused with classical addition and multiplication operators for numbers.
 
\subsection{Qualitative Conjunctive Rule}

The qualitative Conjunctive Rule (qCR)\index{qualitative Conjunctive Rule (qCR)} of $s\geq 2$ sources is defined similarly to the quantitative conjunctive consensus rule, i.e.

\begin{equation}
qm_{qCR}(X)=\sum_{\substack{X_1,\ldots,X_s\in G^\Theta\\ X_1\cap \ldots \cap X_s=X}} \prod_{i=1}^{s} qm_i(X_i)
\label{Chapter1qCR}
\end{equation}

\noindent
The total qualitative conflicting mass is given by $$K_{1\ldots s}=\sum_{\substack{X_1,\ldots,X_s\in G^\Theta\\ X_1\cap \ldots \cap X_s=\emptyset}} \prod_{i=1}^{s} qm_i(X_i)$$

\subsection{Qualitative DSm Classic rule}

The qualitative DSm Classic rule (q-DSmC)\index{qualitative DSm Classic rule (q-DSmC)} for $s\geq 2$  is defined similarly to DSm Classic fusion rule (DSmC) as follows : $qm_{qDSmC}(\emptyset)=L_0$ and for all $X\in D^\Theta\setminus \{\emptyset\}$,

\begin{equation}
qm_{qDSmC}(X)=\sum_{\substack{X_1,,\ldots,X_s\in D^\Theta\\ X_1\cap\ldots \cap X_s=X}} \prod_{i=1}^{s} qm_i(X_i)
\label{Chapter1qDSmC}
\end{equation}

\subsection{Qualitative hybrid DSm rule}

The qualitative hybrid DSm rule (q-DSmH)\index{qualitative hybrid DSm rule (q-DSmH)} is defined similarly to quantitative hybrid DSm rule \cite{Ref-DSmTBook_2004a} as follows: 
\begin{equation}
qm_{qDSmH}(\emptyset)=L_0
\end{equation}
\noindent
and for all $X\in G^\Theta\setminus \{\emptyset\}$
\begin{equation}
qm_{qDSmH}(X)\triangleq
\phi(X)\cdot\Bigl[ qS_1(X) + qS_2(X) + qS_3(X)\Bigr]
 \label{Chapter1qDSmH}
\end{equation}
\noindent
where all sets involved in formulas are in the canonical form and $\phi(X)$ is the {\it{characteristic non-emptiness function}} of a set $X$, i.e. $\phi(X)= L_{m+1}$ if  $X\notin \boldsymbol{\emptyset}$ and $\phi(X)= L_0$ otherwise, where $\boldsymbol{\emptyset}\triangleq\{\boldsymbol{\emptyset}_{\mathcal{M}},\emptyset\}$. $\boldsymbol{\emptyset}_{\mathcal{M}}$ is the set  of all elements of $D^\Theta$ which have been forced to be empty through the constraints of the model $\mathcal{M}$ and $\emptyset$ is the classical/universal empty set. $qS_1(X)\equiv qm_{qDSmC}(X)$, $qS_2(X)$, $qS_3(X)$ are defined by
\begin{equation}
qS_1(X)\triangleq \sum_{\substack{X_1,X_2,\ldots,X_s\in D^\Theta\\ X_1\cap X_2\cap\ldots\cap X_s=X}} \prod_{i=1}^{s} qm_i(X_i)
\end{equation}
\begin{equation}
qS_2(X)\triangleq \sum_{\substack{X_1,X_2,\ldots,X_s\in\boldsymbol{\emptyset}\\ [\mathcal{U}=X]\vee [(\mathcal{U}\in\boldsymbol{\emptyset}) \wedge (X=I_t)]}} \prod_{i=1}^{s} qm_i(X_i)
\end{equation}
\begin{equation}
qS_3(X)\triangleq\sum_{\substack{X_1,X_2,\ldots,X_k\in D^\Theta \\ 
X_1\cup X_2\cup \ldots \cup X_s=X \\ X_1\cap X_2\cap \ldots\cap X_s \in\boldsymbol{\emptyset}}}  \prod_{i=1}^{s} qm_i(X_i)
\end{equation}
\noindent
with $\mathcal{U}\triangleq u(X_1)\cup \ldots \cup u(X_s)$ where $u(X)$ is the union of all $\theta_i$ that compose $X$, $I_t \triangleq \theta_1\cup \ldots\cup \theta_n$ is the total ignorance. $qS_1(X)$ is nothing but the qDSmC rule for $s$ independent sources based on $\mathcal{M}^f(\Theta)$; $qS_2(X)$ is the qualitative mass\index{qualitative mass} of all relatively and absolutely empty sets which is transferred to the total or relative ignorances associated with non existential constraints (if any, like in some dynamic problems); $qS_3(X)$ transfers the sum of relatively empty sets directly onto the canonical disjunctive form of non-empty sets. qDSmH generalizes qDSmC works for any models (free DSm model, Shafer's model or any hybrid models) when manipulating qualitative belief assignments\index{qualitative belief assignment}.

\subsection{Qualitative PCR5 rule (qPCR5)\index{qualitative PCR5 rule (q-PCR5)}}

In classical (i.e. quantitative) DSmT\index{Dezert-Smarandache Theory (DSmT)} framework, the Proportional Conflict Redistribution rule no.~5 (PCR5) defined in \cite{Ref-Book_2006} has been proven to provide very good and coherent results for combining (quantitative) belief masses, see \cite{Ref-Smarandache_2005c,Ref-Dezert_2006}. When dealing with qualitative beliefs within the DSm Field and Linear Algebra of Refined Labels  \cite{Ref-Book_2009} we get an exact qualitative result no matter what fusion rule is used (DSm fusion rules, Dempster's rule, Smets's rule, Dubois-Prade's rule, etc.). The exact qualitative result will a refined label (but the user can round it up or down to the closest integer index label).

\subsection{A simple example of qualitative fusion of qba's}

Let's consider the following set of ordered linguistic labels\index{linguistic labels} $$L=\{L_0,L_1,L_2,L_3,L_4,L_{5}\}$$ 
\noindent
(for example, $L_1$, $L_2$, $L_3$ and $L_4$ may represent the values: $L_1\triangleq \text{{\it{very poor}}}$, $L_2\triangleq \text{{\it{poor}}}$, $L_3\triangleq \text{{\it{good}}}$ and $L_4\triangleq \text{{\it{very good}}}$, where $\triangleq$ symbol means  {\it{by definition}}).\\

Let's consider now a simple two-source case with a 2D frame $\Theta=\{\theta_1,\theta_2\}$, Shafer's model for $\Theta$, and qba's expressed as follows:
$$qm_1(\theta_1)=L_1, \quad qm_1(\theta_2)=L_3, \quad qm_1(\theta_1\cup\theta_2)=L_1$$
$$qm_2(\theta_1)=L_2, \quad qm_2(\theta_2)=L_1, \quad qm_2(\theta_1\cup\theta_2)=L_2$$

\noindent
The two qualitative masses $qm_{1}(.)$ and $qm_{2}(.)$ are normalized since:
$$qm_{1}(\theta_{1})+qm_{1}(\theta_{2})+qm_{1}(\theta_{1}\cup \theta_{2}) = L_{1}+L_{3}+L_{1}= L_{1+3+1}=L_{5}$$
\noindent
and
$$qm_{2}(\theta_{1})+qm_{2}(\theta_{2})+qm_{2}(\theta_{1}\cup \theta_{2}) = L_{2}+L_{1}+L_{2}= L_{2+1+2}=L_{5}$$
We first derive the result of the conjunctive consensus. This yields:
\begin{align*}
qm_{12}(\theta_{1})&=qm_{1}(\theta_{1})qm_{2}(\theta_{1})+qm_{1}(\theta_{1})qm_{2}(\theta_{1}\cup \theta_{2}) + qm_{1}(\theta_{1}\cup \theta_{2})qm_{2}(\theta_{1})\\
&= L_{1}\times L_{2} +L_{1}\times L_{2} +L_{1}\times L_{2}\\
&= L_{\frac{1\cdot 2}{5}} + L_{\frac{1\cdot 2}{5}} +L_{\frac{1\cdot 2}{5}} =L_{\frac{2}{5}+\frac{2}{5}+\frac{2}{5}}=L_{\frac{6}{5}}=L_{1.2}
\end{align*}
\begin{align*}
qm_{12}(\theta_{2})&=qm_{1}(\theta_{2})qm_{2}(\theta_{2})+qm_{1}(\theta_{2})qm_{2}(\theta_{1}\cup \theta_{2}) + qm_{1}(\theta_{1}\cup \theta_{2})qm_{2}(\theta_{2})\\
&= L_{3}\times L_{1} +L_{3}\times L_{2} +L_{1}\times L_{1}\\
&= L_{\frac{3\cdot 1}{5}} + L_{\frac{3\cdot 2}{5}} +L_{\frac{1\cdot 1}{5}} =L_{\frac{3}{5}+\frac{6}{5}+\frac{1}{5}}=L_{\frac{10}{5}}=L_{2}
\end{align*}
\begin{align*}
qm_{12}(\theta_{1}\cup \theta_{2})&=qm_{1}(\theta_{1}\cup \theta_{2}) qm_{2}(\theta_{1}\cup \theta_{2}) = L_{1}\times L_{2}= L_{\frac{1\cdot 2}{5}} =L_{\frac{2}{5}}=L_{0.4}
\end{align*}
\begin{align*}
qm_{12}(\theta_{1}\cap \theta_{2})&=qm_{1}(\theta_{1}) qm_{2}(\theta_{2}) + qm_{1}(\theta_{2}) qm_{2}(\theta_{1}) \\
& = L_{1}\times L_{1} + L_{2}\times L_{3} = L_{\frac{1\cdot 1}{5}} + L_{\frac{2\cdot 3}{5}} \\
& =L_{\frac{1}{5} + \frac{6}{5}}=L_{\frac{7}{5}}= L_{1.4}
\end{align*}

\noindent
Therefore we get:
\begin{itemize}
\item for the fusion with qDSmC, when assuming $\theta_{1}\cap \theta_{2}\neq \emptyset$,
$$qm_{qDSmC}(\theta_{1})=L_{1.2}\qquad qm_{qDSmC}(\theta_{2})=L_{2}$$
$$qm_{qDSmC}(\theta_{1}\cup\theta_{2})=L_{0.4}\qquad qm_{qDSmC}(\theta_{1}\cap\theta_{2})=L_{1.4}$$
\item for the fusion with qDSmH, when assuming $\theta_{1}\cap \theta_{2}= \emptyset$. The mass of $\theta_{1}\cap\theta_{2}$ is transferred to $\theta_{1}\cup\theta_{2}$. 
Hence:
$$qm_{qDSmH}(\theta_{1})=L_{1.2} \qquad qm_{qDSmH}(\theta_{2})=L_{2}$$
$$qm_{qDSmH}(\theta_{1}\cap\theta_{2})=L_{0} \quad qm_{qDSmH}(\theta_{1}\cup\theta_{2})=L_{0.4}+ L_{1.4}=L_{1.8}$$
\item for the fusion with qPCR5, when assuming $\theta_{1}\cap \theta_{2}= \emptyset$. The mass $qm_{12}(\theta_{1}\cap\theta_{2})=L_{1.4}$ is transferred to $\theta_{1}$ and to $\theta_{2}$ in the following way:
$$qm_{12}(\theta_{1}\cap\theta_{2})=qm_{1}(\theta_{1})qm_{2}(\theta_{2}) + qm_{2}(\theta_{1})qm_{1}(\theta_{2})$$
Then, $qm_{1}(\theta_{1})qm_{2}(\theta_{2})=L_{1}\times L_{1}=L_{\frac{1\cdot 1}{5}}=L_{\frac{1}{5}}=L_{0.2}$ is redistributed to $\theta_{1}$ and $\theta_{2}$ proportionally with respect to their qualitative masses put in the conflict $L_{1}$ and respectively $L_{1}$:
$$\frac{x_{\theta_{1}}}{L_{1}}=\frac{y_{\theta_{2}}}{L_{1}}=\frac{L_{0.2}}{L_{1}+L_{1}}=\frac{L_{0.2}}{L_{1+1}}=\frac{L_{0.2}}{L_{2}}=L_{\frac{0.2}{2}\cdot 5}= L_{\frac{1}{2}}=L_{0.5}$$
\noindent
whence $x_{\theta_{1}}=y_{\theta_{2}}=L_{1}\times L_{0.5}=L_{\frac{1\cdot 0.5}{5}}=L_{\frac{0.5}{5}}= L_{0.1}$.

Actually, we could easier see that $qm_{1}(\theta_{1})qm_{2}(\theta_{2})=L_{0.2}$ had in this case to be equally split between $\theta_{1}$ and $\theta_{2}$ since the mass put in the conflict by $\theta_{1}$ and $\theta_{2}$ was the same for each of them: $L_{1}$. Therefore $\frac{L_{0.2}}{2}=L_{\frac{0.2}{2}}=L_{0.1}$.

Similarly, $qm_{2}(\theta_{1})qm_{1}(\theta_{2})=L_{2}\times L_{3}=L_{\frac{2\cdot 3}{5}}=L_{\frac{6}{5}}=L_{1.2}$ has to be redistributed to $\theta_{1}$ and $\theta_{2}$ proportionally with $L_{2}$ and $L_{3}$ respectively :
$$\frac{x'_{\theta_{1}}}{L_{2}}=\frac{y'_{\theta_{2}}}{L_{3}}=\frac{L_{1.2}}{L_{2}+L_{3}}=\frac{L_{1.2}}{L_{2+3}}=\frac{L_{1.2}}{L_{5}}=L_{\frac{1.2}{5}\cdot 5}=L_{1.2}$$
\noindent
whence 
$
\begin{cases}
x'_{\theta_{1}}=L_{2}\times L_{1.2}=L_{\frac{2\cdot 1.2}{5}}=L_{\frac{2.4}{5}}= L_{0.48}\\
y'_{\theta_{2}}=L_{3}\times L_{1.2}=L_{\frac{3\cdot 1.2}{5}}=L_{\frac{3.6}{5}}= L_{0.72}
\end{cases}
$
\noindent
Now, add all these to the qualitative masses of $\theta_{1}$ and $\theta_{2}$ respectively:
\end{itemize}
$$qm_{qPCR5}(\theta_{1})=qm_{12}(\theta_{1}) + x_{\theta_{1}} + x'_{\theta_{1}}=L_{1.2}+L_{0.1}+L_{0.48}=L_{1.2+0.1+0.48}=L_{1.78}$$
$$qm_{qPCR5}(\theta_{2})=qm_{12}(\theta_{2}) + y_{\theta_{2}} + y'_{\theta_{2}}=L_{2}+L_{0.1}+L_{0.72}=L_{2+0.1+0.72}=L_{2.82}$$
$$qm_{qPCR5}(\theta_{1}\cup \theta_{2})=qm_{12}(\theta_{1}\cup \theta_{2})=L_{0.4}$$
$$qm_{qPCR5}(\theta_{1}\cap \theta_{2})=L_{0}$$

The qualitative mass results using all fusion rules (qDSmC,qDSmH,qPCR5) remain normalized in FLARL.\\

Naturally, if one prefers to express the final results with qualitative labels belonging in the original discrete set of labels $L=\{L_0,L_1,L_2,L_3,L_4,L_{5}\}$, some approximations will be necessary to round continuous indexed labels to their closest integer/discrete index value; by example, $qm_{qPCR5}(\theta_{1})=L_{1.78}\approx L_{2}$, $qm_{qPCR5}(\theta_{2})=L_{2.82}\approx L_{3}$ and $qm_{qPCR5}(\theta_{1}\cup \theta_{2})=L_{0.4}\approx L_{0}$.

\subsection{A simple example for the qDSmP transformation}

We first recall that the qualitative extension of 
\eqref{Chapter1eq:DSmP}, denoted $qDSmP_{\epsilon}(.)$ is given by $qDSmP_{\epsilon}(\emptyset)=0$ and $\forall X \in G^\Theta \setminus \{\emptyset\}$ by
\begin{equation}
qDSmP_{\epsilon}(X)=\sum_{Y \in G^\Theta}  \frac{\displaystyle\sum_{\substack{Z \subseteq X\cap Y \\ \mathcal{C}(Z)=1}} qm(Z) + \epsilon\cdot \mathcal{C}(X\cap Y)}{\displaystyle\sum_{\substack{Z \subseteq Y \\ \mathcal{C}(Z)=1}} qm(Z) + \epsilon\cdot \mathcal{C}(Y)}qm(Y)
\label{eq:qDSmP}
\end{equation}
\noindent
where all operations in \eqref{eq:qDSmP} are referred to labels, that is $q$-operators on linguistic labels and not classical operators on numbers.\\

Let's consider the simple frame $\Theta=\{\theta_1,\theta_2\}$ (here $n=\vert\Theta\vert =2$) with Shafer's model (i.e. $\theta_1\cap\theta_2=\emptyset$) and the following set of linguistic labels $L=\{L_0,L_1,L_2,L_3,L_4,L_5\}$, with $L_0=L_{\min}$ and $L_5=L_{\max}=L_{m+1}$ (here $m=4$) and the following qualitative belief assignment: $qm(\theta_1)=L_1$, $qm(\theta_2)=L_3$ and $qm(\theta_1\cup\theta_2)=L_1$. $qm(.)$ is quasi-normalized since $\sum_{X\in 2^\Theta} qm(X)=L_5=L_{\max}$. In this example and with $DSmP$ transformation, $qm(\theta_1\cup\theta_2)=L_1$ is redistributed to  $\theta_1$ and $\theta_2$ proportionally with respect to their qualitative masses $L_1$ and $L_3$ respectively. Since both 
$L_1$ and $L_3$ are different from $L_0$, we can take the tuning parameter $\epsilon=0$ for the best transfer. $\epsilon$ is taken different from zero when a mass of a set involved in a partial or total ignorance is zero (for qualitative masses, it means $L_0$). 

\noindent
Therefore using \eqref{Ref-eq-LabelDivision}, one has
$$\frac{x_{\theta_1}}{L_1}=\frac{x_{\theta_2}}{L_3}=\frac{L_1}{L_1+L_3}=\frac{L_1}{L_4}=L_{\frac{1}{4}\cdot 5}=L_{\frac{5}{4}}=L_{1.25}$$
\noindent
and thus using \eqref{Ref-eq-LabelMultiplication}, one gets
$$x_{\theta_1}=L_1\times L_{1.25}=L_{\frac{1\cdot (1.25)}{5}}=L_{\frac{1.25}{5}}=L_{0.25}$$
$$x_{\theta_2}=L_3\times L_{1.25}=L_{\frac{3\cdot (1.25)}{5}}=L_{\frac{3.75}{5}}=L_{0.75}$$

\noindent
Whence

$$qDSmP_{\epsilon=0}(\theta_1\cap\theta_2)=qDSmP_{\epsilon=0}(\emptyset)=L_0$$
$$qDSmP_{\epsilon=0}(\theta_1)=L_1+ x_{\theta_1}=L_1+ L_{0.25}=L_{1.25}$$
$$qDSmP_{\epsilon=0}(\theta_2)=L_3+ x_{\theta_2}=L_3+ L_{0.75}=L_{3.75}$$

\noindent
Naturally in our example, one has also

\begin{align*}
qDSmP_{\epsilon=0}(\theta_1\cup\theta_2)& =qDSmP_{\epsilon=0}(\theta_1)+qDSmP_{\epsilon=0}(\theta_2) - qDSmP_{\epsilon=0}(\theta_1\cap\theta_2)\\
& =L_{1.25}+L_{3.75} - L_{0}=L_{5}=L_{\max}
\end{align*}

Since $H_{\max}=\log_2 n=\log_2 2=1$, using the qualitative extension of PIC formula \eqref{Chapter1PIC}, one obtains the following qualitative PIC value: 

\begin{align*}
PIC & =1 + \frac{1}{1} \cdot [qDSmP_{\epsilon=0}(\theta_1)\log_2 (qDSmP_{\epsilon=0}(\theta_1)) \\
& \qquad  \qquad \qquad \qquad \qquad + 
qDSmP_{\epsilon=0}(\theta_2)\log_2 (qDSmP_{\epsilon=0}(\theta_2))]\\
& = 1 + L_{1.25} \log_2 (L_{1.25}) + L_{3.75} \log_2 (L_{3.75}) \approx L_{0.94}
\end{align*}
\noindent
since we considered the isomorphic transformation $L_i=i/(m+1)$ (in our particular example $m=4$ interior labels).

\clearpage
\newpage

\section{Belief Conditioning Rules}

\subsection{Shafer's Conditioning Rule (SCR)}

Until very recently, the most commonly used conditioning rule for belief revision was the one proposed by Shafer \cite{Ref-Shafer_1976} and referred here as Shafer's Conditioning Rule (SCR). The SCR consists in combining the prior bba $m(.)$ with a specific bba focused on $A$ with Dempster's rule of combination for transferring the conflicting mass to non-empty sets in order to provide the revised bba. In other words, the conditioning by a proposition $A$, is obtained by SCR as follows :

\begin{equation}
m_{SCR}(.|A)=[m\oplus m_S] (.)
\label{eqSCR}
\end{equation}

\noindent
where $m(.)$ is the prior bba to update, $A$ is the conditioning event, $m_S(.)$ is the bba focused on $A$ defined by $m_S(A)=1$ and $m_S(X)=0$ for all $X\neq A$ and $\oplus$ denotes Dempster's rule of combination \cite{Ref-Shafer_1976}.\\

The SCR approach based on Dempster's rule of combination of the prior bba with the bba focused on the conditioning event remains {\it{subjective}} since actually in such belief revision process both sources are subjective and SCR doesn't manage properly the objective nature/absolute truth carried by the conditioning term. Indeed, when conditioning a prior mass $m(.)$, {\it{knowing}} (or assuming) that the truth is in $A$, means that we have in hands an absolute (not subjective) knowledge, i.e. the truth in $A$ has occurred (or is assumed to have occurred), thus $A$ is realized (or is assumed to be realized) and this is (or at least must be interpreted as) an absolute truth. The conditioning term "Given $A$" must therefore be considered as an absolute truth, while $m_S(A)=1$ introduced in SCR cannot refer to an absolute truth actually, but only to a {\it{subjective certainty}} on the possible occurrence of $A$ from a {\it{virtual}} second source of evidence.  The advantage of SCR remains undoubtedly in its simplicity and the main argument in its favor is its coherence with conditional probability when manipulating Bayesian belief assignment. But in our opinion, SCR should better be interpreted as the fusion of $m(.)$ with a particular subjective bba $m_S(A)=1$ rather than an objective belief conditioning rule. This fundamental remark motivated us to develop a new family of BCR \cite{Ref-Book_2006} based on hyper-power set decomposition (HPSD) explained briefly in the next section. It turns out that many BCR are possible because the redistribution of masses of elements outside of $A$ (the conditioning event) to those inside $A$ can be done in $n$-ways.  This will be briefly presented right after the next section.

\subsection{Hyper-Power Set Secomposition (HPSD)}

Let $\Theta=\{\theta_1,\theta_2,\ldots,\theta_n\}$, $n\geq 2$, a model $\mathcal{M}(\Theta)$ associated for $\Theta$ (free DSm model, hybrid or Shafer's model) and its corresponding hyper-power set $D^\Theta$. Let's consider a (quantitative) basic belief assignment (bba) $m(.): D^\Theta \mapsto [0,1]$ such that $\sum_{X\in D^\Theta}m(X)=1$. Suppose one finds out that the truth is in the set $A\in D^\Theta\setminus\{\emptyset\}$. Let $\mathcal{P}_{\mathcal{D}}(A)=2^A \cap D^{\Theta} \setminus \{\emptyset\}$, i.e. all non-empty parts (subsets) of $A$ which are included in $D^\Theta$. Let's consider the normal  cases when $A\neq\emptyset$ and $\sum_{Y\in \mathcal{P}_{\mathcal{D}}(A)}m(Y)> 0$. For the degenerate case when the truth is in $A=\emptyset$, we consider Smets' open-world, which means that there are other hypotheses $\Theta'=\{\theta_{n+1},\theta_{n+2},\ldots\theta_{n+m}\}$, $m\geq 1$, and the truth is in $A\in D^{\Theta'}\setminus\{\emptyset\}$. If $A=\emptyset$ and we consider a close-world, then it means that the problem is impossible. For another degenerate case, when $\sum_{Y\in \mathcal{P}_{\mathcal{D}}(A)}m(Y)=0$, i.e. when the source gave us a totally (100\%) wrong information $m(.)$, then, we define: $m(A|A)\triangleq 1$ and, as a consequence, $m(X|A)=0$ for any $X\neq A$. Let $s(A)=\{\theta_{i_1},\theta_{i_2},\ldots,\theta_{i_p}\}$, $1\leq p\leq n$,  be the singletons/atoms that compose $A$ (for example, if $A=\theta_1\cup(\theta_3\cap\theta_4)$ then $s(A)=\{\theta_1,\theta_3,\theta_4\}$). The Hyper-Power Set Decomposition (HPSD) of $D^\Theta \setminus \emptyset$ consists in its decomposition into the three following subsets generated by $A$:
\begin{itemize}
\item
$D_1=\mathcal{P}_{\mathcal{D}}(A)$, the parts of $A$ which are included in the hyper-power set, except the empty set; 
\item
$D_2=\{(\Theta\setminus s(A)),\cup , \cap\} \setminus \{\emptyset\}$, i.e. the sub-hyper-power set generated by $\Theta\setminus s(A)$ under $\cup$ and $\cap$, without the empty set.
\item
$D_3=(D^\Theta\setminus\{\emptyset\}) \setminus (D_1\cup D_2)$; each set from $D_3$ has in its formula singletons from both $s(A)$ and $\Theta\setminus s(A)$ in the case when $\Theta\setminus s(A)$ 
is different from empty set.
\end{itemize}
\noindent
$D_1$, $D_2$ and $D_3$ have no element in common two by two and their union is $D^\Theta\setminus\{\emptyset\}$.\\

\noindent
{\it{Simple example of HPSD}}: Let's consider $\Theta=\{\theta_1, \theta_2, \theta_3\}$ with Shafer's model (i.e. all elements of $\Theta$ are exclusive) and let's assume that the truth is in $\theta_2\cup \theta_3$, i.e. the conditioning term is $\theta_2\cup \theta_3$. Then one has the following HPSD: $D_1=\{\theta_2,\theta_3,\theta_2\cup \theta_3\}$, $D_2=\{\theta_1\}$ and $D_3=\{\theta_1\cup \theta_2, \theta_1\cup \theta_3, \theta_1\cup \theta_2\cup \theta_3\}$. More complex and detailed examples can be found in \cite{Ref-Book_2006}.

\subsection{Quantitative belief conditioning rules (BCR)}

Since there exists actually many ways for redistributing the masses of elements outside of $A$ (the conditioning event) to those inside $A$, several BCR's have been proposed in \cite{Ref-Book_2006}. In this introduction, we will not browse all the possibilities for doing these redistributions and all BCR's formulas but only one, the BCR number 17 (i.e. BCR17) which does in our opinion the most refined redistribution since:
\newline - the mass $m(W)$ of each element $W$ in $D_2\cup D_3$ is transferred to those $X\in D_1$ elements which are
included in $W$ if any proportionally with respect to their non-empty masses;
\newline - if no such $X$ exists, the mass $m(W)$ is transferred in a pessimistic/prudent way to the $k$-largest element from $D_1$ which are included in $W$ (in equal parts) if any;
\newline - if neither this way is possible, then $m(W)$ is indiscriminately distributed to all $X \in D_1$
proportionally with respect to their nonzero masses.\\

BCR17 is defined by the following formula (see \cite{Ref-Book_2006}, Chap. 9 for detailed explanations and examples):

\begin{equation}
m_{BCR17}(X|A)=
m(X)\cdot \Bigg[ S_{D_1}
+ 
\displaystyle\sum_{
\begin{array}{c}
\scriptstyle W\in D_2 \cup D_3\\
\scriptstyle X\subset W\\
\scriptstyle S(W)\neq 0
\end{array}
} 
\frac{m(W)}{S(W)}
\Bigg]
+  \displaystyle\sum_{
\begin{array}{c}
\scriptstyle W\in D_2\cup D_3\\
\scriptstyle X\subset W, \, $X$ \,\text{is $k$-largest}\\
\scriptstyle S(W)=0
\end{array}}
m(W)/k
\label{eq:BCR17}
\end{equation}

\noindent
where "$X\,\text{is $k$-largest}$" means that $X$ is the $k$-largest (with respect to inclusion) set included in $W$ and

$$S(W) \triangleq \sum_{Y\in D_1,Y\subset W} m(Y)$$

$$S_{D_1} \triangleq \frac{ 
\displaystyle\sum_{
\begin{array}{c}
\scriptstyle Z\in D_1,\\
\scriptstyle \text{or}\, Z\in D_2 \,\mid\, \nexists Y\in D_1\, \text{with}\, Y\subset Z
\end{array}
} 
m(Z)}{\sum_{Y\in D_1}m(Y)}$$

\noindent
{\textbf{Note:}} The authors mentioned in an Erratum to the printed version of the second volume of DSmT book series ({\verb+http://fs.gallup.unm.edu//Erratum.pdf+})
and they also corrected the online version of the aforementioned book (see page 240 in  {\verb+http://fs.gallup.unm.edu//DSmT-book2.pdf+}
that all denominators of the BCR's formulas are naturally supposed to be different from zero. Of course, Shafer's conditional rule as stated in Theorem 3.6, page 67 of \cite{Ref-Shafer_1976} does not work when the denominator is zero and that's why Shafer has introduced the condition $Bel(\bar{B})< 1$ (or equivalently $Pl(B)>0$) in his theorem when the conditioning term is $B$.\\

\noindent
{\textbf{A simple example for BCR17}}: Let's consider $\Theta=\{\theta_1, \theta_2, \theta_3\}$ with Shafer's model (i.e. all elements of $\Theta$ are exclusive) and let's assume that the truth is in $\theta_2\cup \theta_3$, i.e. the conditioning term is $A\triangleq \theta_2\cup \theta_3$. Then one has the following HPSD: $$D_1=\{\theta_2,\theta_3,\theta_2\cup \theta_3\}, \qquad D_2=\{\theta_1\}$$ 
$$D_3=\{\theta_1\cup \theta_2, \theta_1\cup \theta_3, \theta_1\cup \theta_2\cup \theta_3\}.$$
\noindent
Let's consider the following prior bba:
$m(\theta_1)=0.2$, $m(\theta_2)=0.1$, $m(\theta_3)=0.2$, $m(\theta_1\cup \theta_2)=0.1$, $m(\theta_2\cup \theta_3)=0.1$ and $m(\theta_1\cup \theta_2\cup \theta_3)=0.3$.\\

With BCR17, for $D_2$, $m(\theta_1)=0.2$ is transferred proportionally to all elements of $D_1$, i.e. $\frac{x_{\theta_2}}{0.1}=\frac{y_{\theta_3}}{0.2}=\frac{z_{\theta_2\cup \theta_3}}{0.1}=\frac{0.2}{0.4}=0.5$ whence the parts of $m(\theta_1)$ redistributed to $\theta_2$, $\theta_3$ and $\theta_2\cup\theta_3$ are respectively $x_{\theta_2}=0.05$, $y_{\theta_3}=0.10$, and $z_{\theta_2\cup \theta_3}=0.05$. For $D_3$, there is actually no need to transfer $m(\theta_1\cup \theta_3)$ because $m(\theta_1\cup \theta_3)=0$ in this example; whereas $m(\theta_1\cup \theta_2)=0.1$ is transferred to $\theta_2$ (no case of $k$-elements herein); $m(\theta_1\cup \theta_2\cup \theta_3)=0.3$ is transferred to $\theta_2$, $\theta_3$ and $\theta_2\cup \theta_3$ proportionally to their corresponding masses: 
$$x_{\theta_2}/{0.1}=y_{\theta_3}/{0.2}=z_{\theta_2\cup \theta_3}/{0.1}={0.3}/{0.4}=0.75$$
\noindent
whence $x_{\theta_2}=0.075$, $y_{\theta_3}=0.15$, and $z_{\theta_2\cup \theta_3}=0.075$. Finally, one gets 
\begin{align*}
& m_{BCR17}(\theta_2|\theta_2\cup \theta_3)=0.10+0.05+0.10+0.075=0.325\\
& m_{BCR17}(\theta_3|\theta_2\cup \theta_3)=0.20+0.10+0.15=0.450\\
& m_{BCR17}(\theta_2\cup \theta_3|\theta_2\cup \theta_3)=0.10+0.05+0.075=0.225
\end{align*}
\noindent
which is different from the result obtained with SCR, since one gets in this example:
\begin{align*}
& m_{SCR}(\theta_2|\theta_2\cup \theta_3)=m_{SCR}(\theta_3|\theta_2\cup \theta_3)=0.25\\
& m_{SCR}(\theta_2\cup \theta_3|\theta_2\cup \theta_3)=0.50
\end{align*}
\noindent
More complex and detailed examples can be found in \cite{Ref-Book_2006}.

\subsection{Qualitative belief conditioning rules}

In this section we present only the qualitative belief conditioning rule no 17 which extends the principles of the previous quantitative rule BCR17 in the qualitative domain using the operators on linguistic labels defined previously. We consider from now on a general frame $\Theta=\{\theta_1,\theta_2,\ldots,\theta_n\}$, a given model $\mathcal{M}(\Theta)$ with its hyper-power set $D^\Theta$ and a given extended ordered set $L$ of qualitative values $L=\{L_0,L_1,L_2,\ldots,L_m,L_{m+1}\}$. The prior qualitative basic belief assignment (qbba) taking its values in $L$ is denoted $qm(.)$. We assume in the sequel that the conditioning event is $A\neq\emptyset$, $A\in D^\Theta$, i.e. the absolute truth is in $A$. The approach we present here is a direct extension of BCR17 using FLARL operators. Such extension can be done with all quantitative BCR's rules proposed in \cite{Ref-Book_2006}, but only QBCR17 is presented here for the sake of space limitations.

\subsubsection{Qualitative Belief Conditioning Rule no 17 (QBCR17)}

Similarly to BCR17, QBCR17 is defined by the following formula:
\begin{equation}
qm_{BCR17}(X|A)=
qm(X)\cdot \Bigg[ qS_{D_1}
+ 
\displaystyle\sum_{
\begin{array}{c}
\scriptstyle W\in D_2 \cup D_3\\
\scriptstyle X\subset W\\
\scriptstyle qS(W)\neq 0
\end{array}
} 
\frac{qm(W)}{qS(W)}
\Bigg]
+  \displaystyle\sum_{
\begin{array}{c}
\scriptstyle W\in D_2\cup D_3\\
\scriptstyle X\subset W, \, $X$ \,\text{is $k$-largest}\\
\scriptstyle qS(W)=0
\end{array}}
qm(W)/k
\label{Ref-eq:qBCR17}
\end{equation}

\noindent
where "$X\,\text{is $k$-largest}$" means that $X$ is the $k$-largest (with respect to inclusion) set included in $W$ and

$$qS(W) \triangleq \sum_{Y\in D_1,Y\subset W} qm(Y)$$

$$S_{D_1} \triangleq \frac{ 
\displaystyle\sum_{
\begin{array}{c}
\scriptstyle Z\in D_1,\\
\scriptstyle \text{or}\, Z\in D_2 \,\mid\, \nexists Y\in D_1\, \text{with}\, Y\subset Z
\end{array}
} 
qm(Z)}{\sum_{Y\in D_1}qm(Y)}$$

Naturally, all operators (summation, product, division, etc) involved in the formula \eqref{Ref-eq:qBCR17} are the operators defined in FLARL working on linguistic labels. It is worth to note that the formula \eqref{Ref-eq:qBCR17} requires also the division of the label $qm(W)$ by a scalar $k$.
This division is defined as follows:\\

\noindent
Let $r \in \mathbb{R}, r \neq 0$. Then the label division by a scalar is defined by
\begin{equation}
\frac{L_a}{r} = L_{a/r}
\label{PaperIntroDSmT-eqLabelDivByScalar}
\end{equation}

\subsubsection{A simple example for QBCR17}

\begin{figure}[!h]
\centering
{\tt \setlength{\unitlength}{1pt}
\begin{picture}(90,90)
\thinlines    
\put(20,60){\circle{40}} 
\put(40,60){\circle{40}} 
\put(40,10){\circle{40}} 
\put(60,35){\circle{40}} 
\put(-5,84){\vector(1,-1){10}}
\put(-13,84){$A$}
\put(64,84){\vector(-1,-1){10}}
\put(65,84){$B$}
\put(75,10){\vector(-1,0){15}}
\put(77,7){$C$}
\put(95,35){\vector(-1,0){15}}
\put(97,34){$D$}
\end{picture}}
\vspace{2mm}
\caption{Venn Diagram for the hybrid model for this example.}
\label{venn1}
 \end{figure}
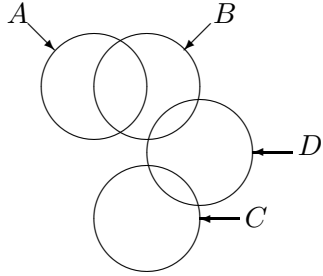

Let's consider $L=\{L_0,L_1,L_2,L_3,L_4,L_5,L_6\}$ a set of ordered linguistic labels\index{linguistic labels}. For example, $L_1$, $L_2$, $L_3$, $L_4$ and $L_5$ may represent the values: $L_1\triangleq \text{{\it{very poor}}}$, $L_2\triangleq \text{{\it{poor}}}$, $L_3\triangleq \text{{\it{medium}}}$, $L_4\triangleq \text{{\it{good}}}$ and $L_5\triangleq \text{{\it{very good}}}$.  Let's consider also the frame $\Theta=\{A,B,C,D\}$ with the hybrid model corresponding to the Venn diagram on Figure \ref{venn1}. 

\noindent
We assume that the prior qualitative bba $qm(.)$ is given by:
$$qm(A)=L_1, \quad qm(C)=L_1, \quad qm(D)=L_4$$

\noindent
and the qualitative masses of all other elements of $G^\Theta$ take the minimal/zero value $L_0$. This qualitative mass is quasi-normalized since $L_1+L_1+L_4=L_{1+1+4}=L_6=L_{\max}$.\\

\noindent If we assume that the conditioning event is the proposition $A\cup B$, i.e. the absolute truth is in $A\cup B$, the hyper-power set decomposition (HPSD) is obtained as follows: $D_1$ is formed by all parts of $A\cup B$, $D_2$ is the set generated by $\{(C,D),\cup,\cap\} \setminus \emptyset=\{C,D,C\cup D, C\cap D\}$, and $D_3=\{A\cup C, A\cup D, B\cup C, B\cup D, A\cup B\cup C, A\cup (C\cap D), \ldots\}$.
Because the truth is in $A\cup B$, $qm(D) = L_4$ is transferred in a prudent way to $(A\cup B)\cap D = B\cap D$ according to our hybrid model, because $B\cap D$ is the 1-largest element from $A\cup B$ which is included in $D$. While $qm(C) = L_1$ is transferred to $A$ only, since it is the only element in $A\cup B$ whose qualitative mass $qm(A)$ is different from $L_{0}$ (zero); hence:
$$qm_{QBCR17}(A) = qm(A) + qm(C) = L_1 + L_1 = L_{1+1} = L_2.$$

\noindent
Therefore, one finally gets:
\begin{align*}
qm_{QBCR17}(A|A\cup B)&=L_2 \qquad & qm_{QBCR17}(C|A\cup B)&=L_0\\
qm_{QBCR17}(D|A\cup B)&=L_0 \qquad & qm_{QBCR17}(B\cap D|A\cup B)&=L_4
\end{align*}
\noindent
which is a normalized qualitative bba.\\

\noindent
More complicated examples based on other QBCR's can be found in \cite{Ref-SmarandacheDezertQBCR2007}.

\section{Conclusion}

A general presentation of the foundations of DSmT has been proposed in this introduction. DSmT proposes new quantitative and qualitative rules of combination for uncertain, imprecise and highly conflicting sources of information. Several applications of DSmT have been proposed recently in the literature and show the potential and the efficiency of this new theory. DSmT offers the possibility to work in different fusion spaces depending on the nature of problem under consideration. Thus, one can work either in $2^\Theta=(\Theta, \cup)$ (i.e. in the classical power set as in DST framework), in $D^\Theta=(\Theta,\cup,\cap)$ (the hyper-power set — also known as Dedekind's lattice) or in the super-power set $S^{\Theta} = (\Theta, \cup, \cap, c(.))$, which includes $2^\Theta$ and $D^\Theta$ and which represents the power set of the minimal refi nement of the frame $\Theta$ when the refinement is possible (because for vague elements whose frontiers are not well known the refinement is not possible). We have enriched the DSmT with a subjective probability ($DSmP_\epsilon$) that gets the best Probabilistic Information Content (PIC) in comparison with other existing subjective probabilities.  Also, we have defined and developed the DSm Field and Linear Algebra of Refined Labels that permit the transformation of any fusion rule to a corresponding qualitative fusion rule which gives an exact qualitative result (i.e. a refined label), so far the best in literature.

\end{document}